\documentclass[10pt,twocolumn,letterpaper]{article}

\usepackage{iccv}              % To produce the CAMERA-READY version
\usepackage{float}

\newcommand{\OurDataset}{Articulate3D}
\newcommand{\OurMethod}{USDNet}

\newcommand{\RevADef}{\textcolor{Green}{R1:\kern+0.3em d2jh}}
\newcommand{\RevBDef}{\textcolor{purple}{R2:\kern+0.3em JwJQ}}
\newcommand{\RevCDef}{\textcolor{brown}{R3:\kern+0.3em vnC8}}
\newcommand{\RevDDef}{\textcolor{blue}{R4:\kern+0.3em p1xj}}
\usepackage{placeins}
\usepackage{makecell}

\definecolor{iccvblue}{rgb}{0.21,0.49,0.74}
\usepackage[pagebackref,breaklinks,colorlinks,allcolors=iccvblue]{hyperref}
\usepackage{subcaption}
\usepackage{float}
\usepackage{graphicx}
\usepackage{booktabs}
\usepackage{xcolor}
\usepackage{comment}
\usepackage{amssymb} % For checkmarks and crosses
\usepackage{multirow}
\usepackage{underscore}
\usepackage{amssymb}% http://ctan.org/pkg/amssymb
\usepackage{pifont}% http://ctan.org/pkg/pifont
\newcommand{\cmark}{\textcolor{LimeGreen}{\normalsize$\checkmark$}}
\newcommand{\checkstar}{\textcolor{Yellow}{\normalsize$\checkmark$}}
\newcommand{\xmark}{\textcolor{RedOrange}{\normalsize$\times$}}
\def\paperID{10170} % *** Enter the Paper ID here
\def\confName{ICCV}
\def\confYear{2025}

\title{Articulate3D: Holistic Understanding of 3D Scenes\\as Universal Scene Description}

\author{Anna-Maria Halacheva$^{1*}$, Yang Miao$^{1*}$, Jan-Nico Zaech$^1$, \\Xi Wang$^{1,2,3}$, Luc Van Gool$^{1}$, Danda Pani Paudel$^1$\\
\\
$^1$INSAIT, Sofia University “St. Kliment Ohridski”,
$^2$ETH Zurich, $^3$TU Munich
}
\begin{document}
% teaser image
\twocolumn[{%
\maketitle
\begin{center}
    \centering
    \vspace{-5pt}\includegraphics[width=\linewidth, trim={0cm 0cm 0cm 0cm}, clip]{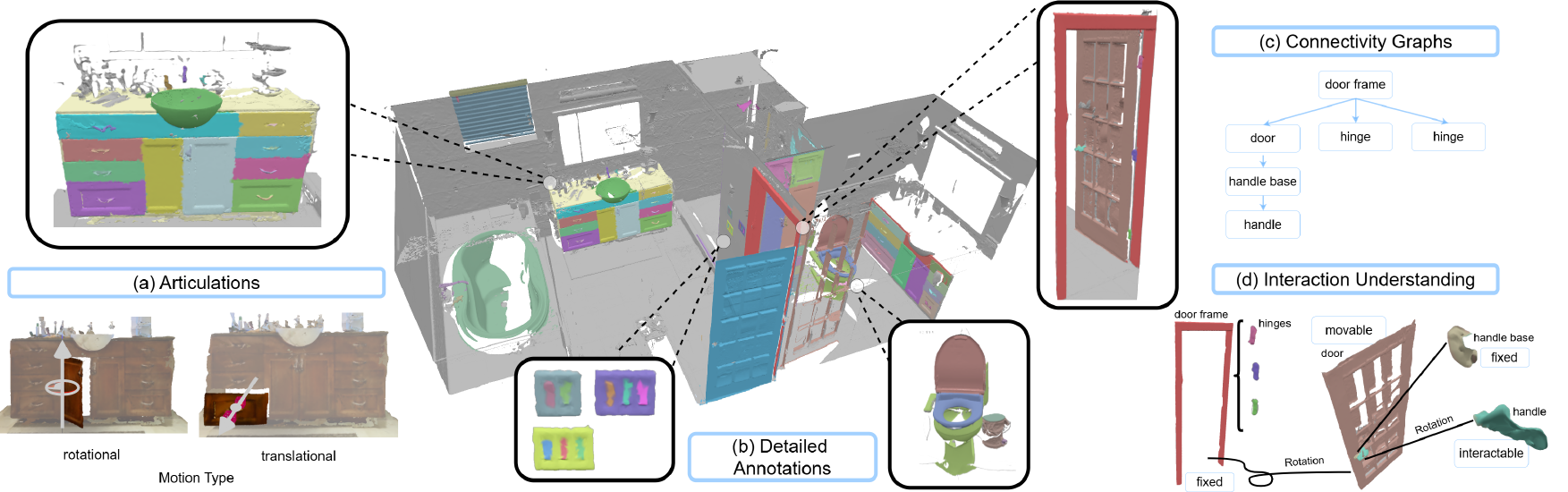}
           \vspace{-15pt}
    \  \captionof{figure}{\textbf{The \protect\OurDataset~Dataset} features (a)  articulation annotations with motion types, origin, axis and range; (b) high-detail segmentations at object and part levels; (c) part connectivity graphs; (d) part articulation roles enabling interaction understanding: movable (articulated) parts, their corresponding interactable parts, and fixed parts.
    %We introduce the first dataset to provide joint annotations on: (1) high-detail semantic segmentation at both the object and part levels, (2) part connectivity graphs, (3) part articulations, and (4) movable (articulated) parts and their corresponding interactable parts, which are used for interactions with the object (e.g., handles). 
   \protect\OurDataset~is the first large-scale real-world dataset of simulation-ready indoor scenes.
    }
    \label{fig:teaser}
\end{center}
}]
% sections
\begin{abstract}
3D scene understanding is a long-standing challenge in computer vision and a key component in enabling mixed reality, wearable computing, and embodied AI. 
Providing a solution to these applications requires a multifaceted approach that covers scene-centric, object-centric, as well as interaction-centric capabilities.
While there exist numerous datasets and algorithms approaching the former two problems, the task of understanding interactable and articulated objects is underrepresented and only partly covered in the research field. 
In this work, we address this shortcoming by introducing: 
(1) \protect\OurDataset{}, an expertly curated 3D dataset featuring high-quality manual annotations on 280 indoor scenes. 
\protect\OurDataset{} provides 8 types of annotations for articulated objects, covering parts and detailed motion information, %-- object and part segmentations, connectivity, movable, interactable and fixed parts, motion parameters, and object mass annotations -- 
all stored in a standardized scene representation format designed for scalable 3D content creation, exchange and seamless integration into simulation environments. 
% the USD format for seamless integration with downstream tasks; 
(2) \protect\OurMethod{}, a novel unified framework capable of simultaneously predicting part segmentation along with a full specification of motion attributes for articulated objects. 
We evaluate \protect\OurMethod{} on \protect\OurDataset{} as well as two existing datasets, demonstrating the advantage of our unified dense prediction approach. 
Furthermore, we highlight the value of \protect\OurDataset{} through cross-dataset and cross-domain evaluations and showcase its applicability in downstream tasks such as scene editing through LLM prompting and robotic policy training for articulated object manipulation. 
We provide \href{https://insait-institute.github.io/articulate3d.github.io/}{open access} to our dataset, benchmark, and method’s source code.
\end{abstract}

\renewcommand{\thefootnote}{\fnsymbol{footnote}} % Use symbols instead of numbers for footnotes
\footnotetext[1]{Equal contribution}
\renewcommand{\thefootnote}{\arabic{footnote}} 
\section{Introduction}
3D holistic scene understanding requires understanding at various semantic and spatial levels, including 3D object detection, semantic/instance segmentation, fine-grained part segmentation, object affordance prediction, and scene hierarchy prediction~\cite{survey3DVQA, multimodel3Dneurips2024}. 
Achieving these capabilities relies on high-quality, richly annotated, large-scale datasets. 
While numerous 3D indoor scene datasets exist, most focus solely on object-level semantic segmentation, lacking part-level or articulation information~\cite{scannet,matterport3d,yeshwanthliu2023scannetpp,Wald2019RIO,SceneNN}.
Among existing datasets, MultiScan \cite{mao2022multiscan} and SceneFun3D \cite{delitzas2024scenefun3d} %are most relevant to our work, 
% and both incorporate articulation annotations with part segmentation at a limited scale~(117/230 scans respectively).
incorporate articulation annotations with part segmentation. % but at a limited scale.
However, both datasets lack annotations for connectivity and multi-granularity hierarchies (e.g., room-object-part relationships), constraining their use in the tasks of holistic scene understanding and embodied AI. 
On the other hand, synthetic environments and auto-generated 3D assets~\cite{sapien, jia2024sceneverse, embodiedscan} are often used to complement the limited scale and coverage of real-world datasets. However, reliance on synthetic data introduces a significant sim-to-real gap~\cite{Villasevil-RSS-24,syntoreal24eccv}, where models trained on synthetic scenes struggle to generalize effectively to real-world environments. 
This gap highlights the need for real-world datasets to ensure effective model adaptation and deployment in practical applications.

To address the limitations of existing datasets, we introduce \protect\OurDataset{}, a large-scale, high-quality and richly annotated 3D dataset in Universal Scene Description (USD) format. 
\protect\OurDataset{} contains 280 scenes of high-resolution scans based on Scannet++~\cite{yeshwanthliu2023scannetpp}.
As illustrated in Fig. \ref{fig:teaser}, it features multi-level and high-quality annotations:
% (a) articulation annotations,
(a) high-detail semantic instance segmentations at both object and part levels,
(b) connectivity graphs linking scene entities,
(c) articulation roles of the parts, enabling in-depth interaction understanding. Specifically, we annotate movable (articulated) parts, their corresponding interactable parts, their corresponding motion parameters, and fixed parts, the latter allowing us to differentiate between freely movable and stationary elements. 
These rich annotations make \protect\OurDataset{} the most comprehensive scene articulation dataset, 
% tailored to support holistic 3D scene understanding, 
as highlighted in Tab. \ref{tab:annotationsglossary}. 
Notably, we integrate our detailed annotations with 3D scans in USD format, facilitating a wide range of applications in 3D computer vision and embodied AI (see demonstrations in Sec.~\ref{sec:applicationmainpaper}).  Given USD's utility, its established importance in robotics \cite{NguyenUSD,Villasevil-RSS-24,Jacinto}, and the difficulties associated with asset USD conversion \cite{Villasevil-RSS-24}, \protect\OurDataset's native USD construction offers substantial practical value.

We also propose \protect\OurMethod{}, a novel method that jointly predicts object parts and articulation from 3D point clouds.
\protect\OurMethod{} utilizes a point-wise dense prediction mechanism, leveraging part-related point features to improve articulation prediction performance.
We evaluate \protect\OurMethod{} on \protect\OurDataset{} as well as two existing datasets across three tasks: movable part segmentation, articulation prediction, and interactable part segmentation. 
The results demonstrate the advantages of our joint prediction framework with improvement of \textbf{5.7}$\%$ for motion parameter prediction,   \textbf{2.7}$\%$ for movable part segmentation and 0.9$\%$ for interactable part segmentation compared to the baselines. 

Furthermore, we highlight the value of \protect\OurDataset{} through cross-dataset and cross-domain evaluations, showing that training on \protect\OurDataset{} improves generalization performance, with improvement of \textbf{5.7}$\%$ in the cross-dataset setting and \textbf{12.4}$\%$ in the cross-domain setting. 
We further demonstrate the applicability of \protect\OurDataset{} in two downstream tasks: (1) 3D scene editing via LLM-based prompting, and (2) robotic policy training for articulated object manipulation, showing that our USD scenes can be directly integrated into simulation environments without requiring manual adaptations (Sec. \ref{sec:applicationmainpaper}). 
Overall, this work makes the following contributions:
\begin{itemize}
    \item We introduce \protect\OurDataset{}, the richest articulation dataset for real-world 3D scenes, combining semantic part and object instance segmentations, connectivity, and detailed articulation annotations. 
    \protect\OurDataset{} is the first large-scale, non-synthetic dataset provided in a simulation-ready USD format, with significantly greater object variability than existing synthetic datasets (Tab. \ref{tab:annotationsglossary}).
    \item We propose \protect\OurMethod{}, a novel method that jointly predicts 3D movable and interactable part segmentation and articulation from 3D scene point clouds. 
    \item We further demonstrate the value of our dataset \protect\OurDataset{} in four different downstream tasks, highlighting its applicability in various scenarios. 
\end{itemize}

\section{Related Work}
We present an overview of related work on datasets for semantic scene understanding and 3D object articulation. We provide summarized overviews on the annotation sets and the diversity within synthetic and real-world datasets in Tab. \ref{tab:annotationsglossary}.  Additionally, we review existing algorithms for holistic scene understanding.

\begin{table}
    \centering
    \scriptsize
    \begin{tabular}{@{\hskip 0.5mm}l|c@{\hskip 0.5mm}c@{\hskip 0.5mm}c@{\hskip 0.5mm}|c@{\hskip 0.5mm}c@{\hskip 0.5mm}c@{\hskip 0.5mm}c@{\hskip 0.5mm}|c@{\hskip 0.5mm}c@{\hskip 1.6mm}|c@{\hskip 1.6mm}c@{\hskip 1.6mm}c@{\hskip 1.6mm}}
        \multirow{2}{*}{Dataset} & \multicolumn{3}{c|}{Instance} & \multicolumn{4}{c|}{Articulation} & \multicolumn{2}{c|}{Scene} & \multicolumn{3}{c}{Metadata} \\  
       & \rotatebox{90}{Object} & \rotatebox{90}{Part} & \rotatebox{90}{Connectivity} & \rotatebox{90}{Motion-Param.} & \rotatebox{90}{Movable} & \rotatebox{90}{Interactable} & \rotatebox{90}{Fixed} & \rotatebox{90}{HD-3D} & \rotatebox{90}{Sim-Ready} & \rotatebox{90}{Scenes} & \rotatebox{90}{Unique Obj.} & \rotatebox{90}{Sem. Labels}  \\  
        \midrule
        \multicolumn{13}{c}{Synthetic}\\
        \midrule
         ProcTHOR \cite{procthor} & \cmark & \checkstar & \cmark & \cmark & \cmark & \checkstar & \xmark & \cmark & \cmark & 10k & 1.6k & 108  \\  
         HSSD-200 \cite{khanna24hssd} & \cmark & \xmark & \xmark & \xmark & \xmark & \xmark & \xmark& \cmark & \cmark & 211 & 18.7k & 466 \\  
        RoboCasa \cite{robocasa2024} & \cmark & \checkstar & \cmark & \cmark & \cmark & \checkstar & \xmark & \cmark & \cmark & 120 & 2.5k & 153  \\  
        \midrule
        \multicolumn{13}{c}{Real-World}\\
        \midrule
         ScanNet \cite{scannet} & \cmark & \xmark & \xmark & \xmark & \xmark & \xmark & \xmark & \xmark & \xmark & 1.5k  & 36k & 1.1k \\  
        Matterport3D \cite{matterport3d} & \cmark & \xmark & \xmark & \xmark & \xmark & \xmark & \xmark & \xmark & \xmark & 90 & 50k & 1.6k  \\  
        3RScan \cite{Wald2019RIO} & \cmark & \xmark & \xmark & \xmark & \xmark & \xmark & \xmark & \xmark & \xmark & 478 & 48k & 534  \\  
        ARKitScenes \cite{baruch2021arkitscenes} & \xmark & \xmark & \xmark & \xmark & \xmark & \xmark & \xmark & \cmark & \xmark & 1.7k & - & -  \\  
        ScanNet++* \cite{yeshwanthliu2023scannetpp} & \cmark & \xmark & \xmark & \xmark & \xmark & \xmark & \xmark & \cmark & \xmark & 380 & 30k & 1000  \\  
        MultiScan \cite{mao2022multiscan} & \cmark & \checkstar & \cmark & \cmark & \checkstar & \xmark & \xmark & \xmark & \xmark & 117 & 11k & 419  \\ 
        SceneFun3D* \cite{delitzas2024scenefun3d} & \xmark & \checkstar & \xmark & \cmark & \checkstar & \cmark & \xmark & \cmark & \xmark & 315 & - & - \\  
         
        \midrule
        \textbf{\protect\OurDataset~ (Ours)} & \cmark & \cmark & \cmark & \cmark & \cmark & \cmark & \cmark & \cmark & \cmark & 280 & 30k & 1000 \\  
    \end{tabular}
    \caption{Comparison of real-world and synthetic indoor scenes datasets.  Articulate3D is the first dataset to combine high-quality real-world scene scans with instance segmentations on both object and part-level, connectivity and all types of articulation annotations necessary to execute/simulate interactions in 3D scenes. Articulate3D features high-definition scenes (HD-3D), provided in the USD format, delivering simulation-ready assets. \protect\OurDataset is also more diverse than synthetic datasets both in terms of unique instances in the scenes and in semantic labels. \\Legend: 
    * Numbers based on released scenes. \checkstar~ Annotations with partial coverage. \cite{robocasa2024,procthor}: motion-linked parts only. \cite{mao2022multiscan}: no movable interactable components. \cite{delitzas2024scenefun3d}: only interactable parts.}
    \label{tab:annotationsglossary}
    \vspace{-0.6cm}
\end{table}

% \subsection{Dataset - 3D Semantic Understanding}
\par\noindent\textbf{Dataset - 3D Semantic Understanding}
% Understanding 3D scenes 
is crucial for various applications, including mixed reality  \cite{zhai2023commonscenes,Zijie24eccv,li2024dreamscene,wang2024roomtex,ocal2024sceneteller}, scene reasoning  \cite{huang2024chatscene,ding2022language}, and embodied AI  \cite{scarpellini24eccv,yang2024holodeck,Prabhudesai20cvpr}. 
Progress in these areas requires high-quality, multimodal 3D data \cite{chen2024x360,OpenEQA2023,multiply}, driving the development of various scene datasets. 
%In response, various 3D scene datasets have been introduced. 
ScanNet++ \cite{yeshwanthliu2023scannetpp}, ScanNet \cite{scannet}, Matterport3D \cite{matterport3d} and others \cite{Wald2019RIO,SceneNN,yadav2022habitat} offer instance segmentation with extensive label details. Another line of work prioritizes high-definition scans and scale, providing very limited to no annotations, e.g. ARKitScenes \cite{baruch2021arkitscenes}, HM3D~\cite{ramakrishnan2021hm3d}. Yet, despite the diversity in available datasets, most provide a single annotation type and offer only object-level semantic segmentation \cite{scannet,yeshwanthliu2023scannetpp,matterport3d,Wald2019RIO,SceneNN,yadav2022habitat}, limiting their applicability for tasks requiring finer detail. Although datasets with both object- and part-level annotations exist \cite{Mo2019CVPR}, those consist of isolated objects rather than complete scenes, reducing their suitability for holistic 3D scene understanding.

\protect\OurDataset~addresses these gaps by providing high-definition 3D scans with label-rich object- and part-level segmentation, as well as connectivity graphs, mass annotations, and extensive articulation data, creating a foundation for advanced 3D scene understanding.

% \subsection{Dataset - 3D Interaction}
\par\noindent\textbf{Dataset - 3D Articulation Information} is essential for 3D scene understanding, enabling realistic object manipulation and interaction. 
%Recent research on learning articulation patterns \cite{song2024reacto,jiayi2023paris,jiang2022ditto}  and representing articulated states \cite{leia24eccv,liu2024cage,Shengyi22cvpr}, shows the need for richly-annotated articulation datasets to support tasks requiring dynamic scene exploration and interaction. 
Articulation learning \cite{song2024reacto,jiayi2023paris,jiang2022ditto} and articulated state representation research \cite{leia24eccv,liu2024cage,Shengyi22cvpr} highlight the need for richly-annotated articulation datasets for dynamic scene tasks.
%Recent datasets have expanded the scope of object articulation annotations, but still lack the full details needed for comprehensive representation, e.g. in a simulation.
Despite recent annotation improvements, current articulation datasets lack the detail for comprehensive representation, e.g., in a simulation. SceneFun3D \cite{delitzas2024scenefun3d} annotates interactable parts and their motion within ARKitScenes, but lacks object instance segmentation, movable parts, and connectivity annotations essential for scene understanding. MultiScan \cite{mao2022multiscan} provides part-level articulations and segmentations but not the interactable components necessary for practical manipulation. Its movability annotations also rely on multiple scans of objects in different states, complicating processing and extensibility. Other datasets offer broader annotations but include only individual objects \cite{Liu2022AKB48AR,Xiang2020SAPIEN}, limiting their use on scene level. 
Our dataset addresses these limitations by providing an extensive set of annotation types, covering movable and interactable parts, and precise movement specifications (Tab. \ref{tab:annotationsglossary}). Using the USD format, we provide simulation-ready scenes without multiple scans, and enable integration into different robotics and embodied AI workflows \cite{Villasevil-RSS-24,NguyenUSD}. 

\paragraph{Dataset - Synthetic and Generated 3D Assets}, as the complement to the challenging to obtain real-world datasets, are receiving significant research attention~\cite{sapien,procthor, jia2024sceneverse, embodiedscan, robocasa2024, Villasevil-RSS-24,syntoreal24eccv, chen2024urdformer}.
Recent datasets primarily focus on either segmentation \cite{khanna24hssd,embodiedscan,Structured3D} or articulation \cite{robocasa2024,procthor,replicaCAD}. While earlier efforts targeted datasets size, e.g., ProcTHOR \cite{procthor} (10,000 scenes), recent trends favor smaller, expertly curated datasets with greater diversity, e.g., RoboCasa \cite{robocasa2024} (120 scenes) and HSSD-200 \cite{khanna24hssd} (211 scenes). \cite{khanna24hssd} further demonstrated that training on 122 high-quality scenes outperforms using the entire ProcTHOR dataset for generalization. Yet, compared to real-world data \cite{yeshwanthliu2023scannetpp}, those datasets still have limited asset variability (Tab. \ref{tab:annotationsglossary}). The synthetic articulation-oriented works also lack annotation categories required for in-depth interaction understanding.

Another line of research focuses on generating synthetic articulated data \cite{chen2024urdformer,gao2024meshart,liu2024cage}. However, these approaches are typically trained exclusively on synthetic 3D assets, resulting in a synthetic-to-real gap that can significantly impact model performance in real-world scenarios \cite{Villasevil-RSS-24,syntoreal24eccv}. \cite{Villasevil-RSS-24} shows that a robotic policy trained on a large scale synthetic data does not generalize on real-world targets, achieving only 10\% success rate. Bridging this gap requires high-quality real-world datasets with comprehensive annotations - one such example being \protect\OurDataset, as shown in Sec. \ref{sec:applicationmainpaper}.

% \subsection{Algorithms - Holistic Scene Understanding}
\par\noindent\textbf{Algorithms - Holistic Scene Understanding} involves recognizing objects, their spatial relationships, articulations and contextual information about the scene as a whole.
We focus on part-level segmentation and articulation within indoor scenes. One popular approach is 2D image-based methods~\cite{opdsingle, opdmulti, detechwhatcan, visaulidenarti, holisticobjectrecog, learnfromsyntheimage} that use RGB(-D) images for 2D/3D part-level scene understanding tasks.
% \cite{learn_from_synthe_image} learns the mapping from 2D images to 3D human articulations with rendered 2D images. 
\cite{opdsingle, opdmulti} address the task of detecting openable part detection and their 3D motion parameters from a single image, while \cite{ detechwhatcan, visaulidenarti, holisticobjectrecog} focus on object-part-level understanding in 2D. 
However, those methods are limited to single objects or small-scale scenes due to the restricted viewpoint of images.
3D-based methods offer an alternative by directly processing 3D data to achieve holistic scene understanding.
Pioneered by \cite{pointnet, sparsecnn, hais}, research in 3D semantic/instance scene understanding has gained popularity~\cite{vu2022softgroup, mask3d, rozenberszki2024unscene3d, pointnet, takmaz2023openmask3d, sgiformer}.
Meanwhile, due to lack of high-quality 3D dataset with complete articulations, 3D holistic scene understanding of articulation is still an open problem. 
Recently, there has been growing interest in developing methods that capture the articulation intricacies~\cite{Shape2Motion, Shi2022P3NetPM, mao2022multiscan, delitzas2024scenefun3d, RPMNet19, Shi2020SelfSupervisedLO}.
However, those methods either focus on single object parsing with synthetic data, or on partial articulation prediction, limiting their applications in real-world scenarios.
To address this, we build a baseline using recent 3D instance segmentation frameworks, such as \cite{mask3d} and \cite{vu2022softgroup}, with verified performance and publicly available code. 

\begin{figure*}
    \centering
    \includegraphics[width=\linewidth]{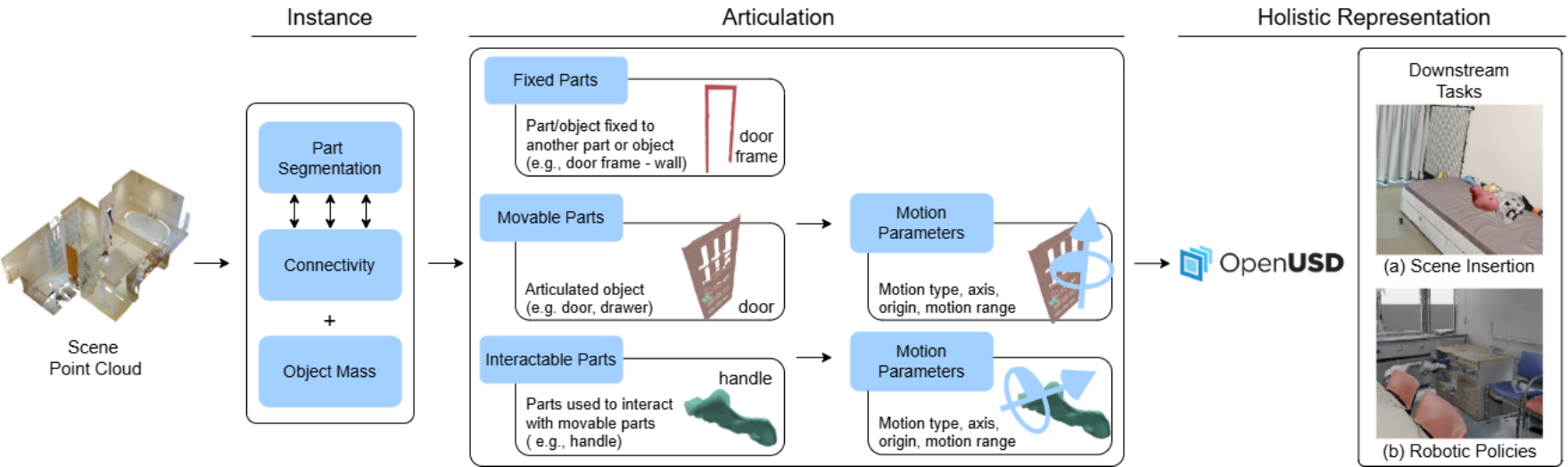}
     \vspace{-15pt}
    \caption{\textbf{Overview of the annotations in \protect\OurDataset}, leading us to the creation of USD scene representations. We determine the objects, their mass and parts, as well as the part connectivity. We annotate motion type (rotational, translational, none), and based on it - the articulations. We identify the role of the different parts within the articulations (movable or interactable). For each articulated part we also capture the motion parameters axis, origin (only for rotation) and motion range, or a fixture point for fixed objects. We use the annotations to define the USD (OpenUSD) scene representation, enabling various downstream tasks, as described in detail in Sec. \ref{sec:applicationmainpaper}.}
    \label{fig:annotations}
    \vspace{-10pt}
\end{figure*}

\section{Building the \protect\OurDataset~Dataset}\label{sec:builddataset}
In this section, we firstly motivate our choice of the USD format for scene representation and then detail the construction of \protect\OurDataset,
% , which builds upon the ScanNet++ dataset \cite{yeshwanthliu2023scannetpp} by adding part-level annotations, articulation information, part connectivity, and object mass annotations, 
which enables a holistic representation in USD (OpenUSD) format, as illustrated in Sec. \ref{fig:annotations}. 
% Our work uses the high-quality data and dense semantic annotations of ScanNet++ as a foundation, upon which we layer additional annotations for enhanced scene understanding and interaction modeling.

% \par\noindent\textbf{USD preliminaries.}
\par\noindent\textbf{USD preliminaries.}
Universal Scene Description~(USD) organizes a scene into hierarchical entities--primitives (prims)--the building blocks representing all objects and relationships in the scene. 
Prims support a nested structure, where complex objects (e.g., cabinets) are represented as parent prims containing child sub-prims.
Each prim can be assigned various attributes, e.g., geometry, pose, scale, appearance, and custom attributes~(physical properties e.g., mass, semantic labels, etc). 
By carefully designing the annotation pipeline and a framework to integrate 3D scenes with annotations, we provide a solution that can automatically transform the whole scene to simulation-ready USD format, addressing the challenges presented in~\cite{Villasevil-RSS-24}.
In summary, \protect\OurDataset{} provides scenes with rich annotations in USD, offering the following advantages: 
1) enabling hierarchical and structured parsing of the scene, 
2) using physics-aware simulation-ready format on realistic scenes and facilitating Sim2Real learning \cite{Villasevil-RSS-24} at scale.  
% 2) enabling transfer of 3D assets to simulation-ready USD, which is reportedly challenging [56]. 
3) making the real-world scenes LLM-interpretable and editable~(Sec.~\ref{sec:applicationmainpaper} \& Sec. \ref{suppsec:downstream}). 
More details on USD can be found in 
 Sec. \ref{suppsec:usd}.
% We moved USD details to the supplementary material, adding a discussion of simulation-based training and ArtMask3D details in its place.
 
\subsection{ScanNet++ as a Foundation}
ScanNet++~\cite{yeshwanthliu2023scannetpp} provides 380 richly annotated 3D scenes captured using multi-source data collection strategy (Faro Focus Premium laser scanner for high-quality point clouds, together with RGB-D streams and high-resolution DSLR images for each scene). 
% This strategy makes the dataset suitable for tasks that integrate 2D and 3D information~\cite{Huang2023Segment3D,takmaz2023openmask3d,Peng2023OpenScene,nguyen2024open3dis}. The accurate alignment between the RGB-D data, DSLR images, and laser-scanned point clouds makes ScanNet++ the ideal choice for building \protect\OurDataset.
% \par
% \noindent\textbf{Choice of ScanNet++}
% We chose ScanNet++ 
Compared to other datasets (e.g., MultiScan \cite{mao2022multiscan}, ScanNet \cite{scannet}, Matterport3D \cite{matterport3d}, Tanks and Temples \cite{tanksTemples}, ETH3D \cite{eth3d}), ScanNet++ provides (1) superior data resolution, (2) extensive laser scan coverage, and (3) dense semantic annotations. Notably, while ARKitScenes \cite{baruch2021arkitscenes} offers comparable mesh quality, it does not provide semantic segmentations, making ScanNet++ the only ideal basis for the complex annotations needed in  \protect\OurDataset.
Our work focuses on the 280 training scenes from ScanNet++, as no object-level semantic annotations are publicly available for the 50 validation and 50 test scenes.
% This subset of fully labeled scenes provides the necessary foundation for detailed part-level, connectivity, and articulation annotations in \protect\OurDataset.

\subsection{Building  \protect\OurDataset} \label
Applications like embodied AI require precise articulation data, and simulation learning relies on annotated physical properties. A shared foundation for many of these tasks is robust semantic instance segmentation. In response to the needs for such data, the \protect\OurDataset~dataset introduces new layers of annotation onto ScanNet++’s base.  Our annotations cover part segmentation, articulation, part connectivity, and physical properties. Fig. \ref{fig:annotations} provides an overview of the different annotation categories and their associated parameters. In addition to annotating the parts motion parameters motion type, axis, origin, and range, we also annotate the interactable parts that indicate potential contact regions, e.g., handles for doors. Furthermore, fixed attachments between parts and objects, as well as the mass of non-fixed objects, are also provided to support physical simulations.

To ensure consistency across scenes, our annotation approach defines a standardized set of 50 object classes (e.g., cabinets, windows, ovens) and their corresponding segmentable parts. We annotate overall 70 part classes. A complete list of the 120 object and part labels, including distribution, is provided in the supplementary material. Within these labels, we allow for some flexibility to capture semantic nuances (e.g., "cabinet" vs. "closet") and functional distinctions (e.g., "faucet control" vs. "faucet handle"), preserving subtle distinctions that may be valuable for downstream tasks. We note that label synonyms are merged (e.g., "trash bin" and "trash can") and counted as a single object class. A list with the used synonyms and label variations will be released within the dataset.

To ensure thorough and accurate annotation, each scene undergoes a two-stage process. The annotation pipeline (Fig. \ref{fig:tool}) is based on MultiScan’s annotation tools with extensions to support connectivity annotation, and articulation suggestions for motion axis, origin and range. 
In the first stage, annotators focus on part segmentation and connectivity annotations. 
The second stage focuses on articulation annotation, capturing part roles (movable, interactable, fixed) and motion parameter attributes (motion type, axis, origin, range), as shown in  Fig. \ref{fig:annotations}.
% Our team consists of five expert annotators who conduct the primary annotations, followed by a sixth expert who reviews and refines all submitted annotations. This quality control step is critical in maintaining consistency and accuracy across the dataset.
To maintain dataset consistency and accuracy, we employ a two-tiered annotation process. Five expert annotators conduct primary annotations, which are then reviewed and refined by a sixth expert.

\par\noindent\textbf{Part Segmentation and Connectivity}
% Our part segmentation process involves manual annotation by expert annotators, who 
%
Our expert annotators manually segment, label, and connect parts within the predefined object set. To enhance accuracy, we first use Felzenszwalb and Huttenlocher’s hierarchical graph-based segmentation algorithm \cite{Felzenszwalb} and provide additional oversegmentation of the objects to annotators.
% with ScanNet++ object class labels relevant to our annotations (e.g., doors, windows, cabinets, fridges, ovens). To obtain this oversegmentation, we use Felzenszwalb and Huttenlocher’s hierarchical graph-based segmentation algorithm \cite{Felzenszwalb}.
The oversegmentation divides the object mesh into smaller, more manageable segments, with the coarse level capturing larger parts and the fine level delineating finer boundaries at the level of individual triangles. This enables annotators to use these pre-segmented boundaries as a starting point, significantly speeding up the annotation process. Where appropriate, annotators can refine these segments to capture finer details or correct inaccuracies in the automated segmentation (Fig. \ref{fig:overseg}). The fine-grained mesh-level labeling also enables adjustments addressing inaccuracies present in ScanNet++'s original instance segmentation.

Part connectivity is defined through hierarchical connectivity graphs, with the "root" node being the base and "child" nodes representing physically attached (e.g., door to cabinet) or functionally associated (e.g., lid to container) components.
To ensure the quality of the connectivity annotations, we implement automated checks to detect potential issues. These checks identify instances of double-parenting and gaps in the hierarchy, alerting annotators to inconsistencies or errors. By combining expert review with automated validation, we ensure that even complex annotations meet high standards of accuracy and consistency, resulting in a dataset that is both detailed and highly accurate.

\par\noindent\textbf{Articulation Annotation}
For each part we annotate its role within its articulation structure - movable, interactable, with a fixed attachment, or none. For the movable parts, the annotators label the base part (the stationary reference), and the motion parameters motion type (either translational or rotational), axis, origin, and  range (angle limits for rotation or distance limits for translation), as shown in Fig. \ref{fig:annotations}.

To speed up the process, we provide annotators with semi-automated suggestions. For hinge-based rotations (e.g., doors and windows), we estimate the rotation axis along the longest vertical edge of the bounding box. For sliding parts (e.g., drawers), we use surface normals from the object’s front face to predict the translation axis. These suggestions are then refined by annotators. 
Fixed articulations are also introduced to represent immovable fixed parts, such as ceiling lights attached to ceilings or door frames anchored to walls. To determine which objects and parts can be fixed, we define a set of eligible class labels. Within each scene, candidates for a fixture are proposed based on the proximity of bounding boxes and an example attachment point of closest proximity to the two objects is determined. Annotators then verify and refine the proposals.
% , ensuring accurate and consistent labeling.

\par\noindent\textbf{Mass Annotation}
We assign a predefined mass to each object class, determined by a GPT-4o \cite{gpt4o} estimation of the typical object's weight within that class. Instance-level mass estimation is performed by scaling the class mass based on individual bounding box volumes. The average bounding box size is assigned to the predefined weight and other instances are scaled accordingly. By providing instance-specific weight estimates, we enable realistic physical simulations and improved scene understanding.

\subsection{\protect\OurDataset~Statistics}
Our dataset includes annotations on 8 different categories - object and part segmentations, movable,  interactable and fixed parts, motion parameters, connectivity, and object mass annotations.
The dataset includes over 15k segments of 120 class labels - 50 object classes and 70 part classes. 
On average, each scene has 45 segmentations and 11 connection graphs, with an average hierarchy depth of 4, including the root part. 
For part annotations, $64\%$ of parts are articulated, with
$33\%/35\%$ interactable in the train/test set respectively. 
Among articulated part segments, $25\%/26\%$ are of translations in train/test respectively.
Our segments on average cover 25\% of the mesh vertices in each scene, excluding the background elements such as the floor, walls, and ceiling. 
% On the articulation level, \protect\OurDataset~includes 21 articulations per scene on average, with $\approx$70\% being rotational
Additional label lists, lists of synonyms within the dataset (not counted in the 120 labels), and information on label distribution are provided in Sec. \ref{suppsec:datasetstatistics}.

\begin{figure*}
\begin{center}
\includegraphics[width=0.99\linewidth]{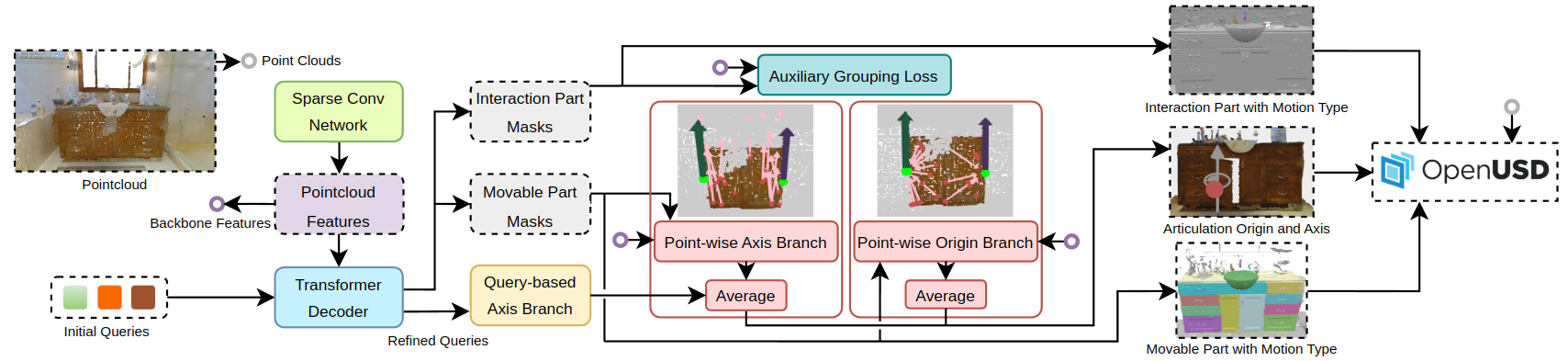}
      \vspace{-5pt}
    \caption{Framework of \protect\OurMethod~for movable/interactable part segmentation, motion type and articulation prediction. The boarders of data and variables are in dashed lines. }   \label{fig:usdnet}
\end{center}
\vspace{-22pt}
\end{figure*}

\section{\protect\OurMethod}\label{sec:method}
% \todo{write a bit about the goal of having a unified framework for predicting all the elements in USD; also motivation for our design}
We propose \protect\OurMethod{}, a unified framework of simultaneously predicting all necessary components for the construction of an interaction-oriented USD~(movable and interactable parts, and motion parameters) given the point cloud of a scene.
Given a scene point cloud as input, \protect\OurMethod{} predicts 
(1) instance masks $M=\{m_k\}^K_{k=1}$ of movable parts  and their motion type $C_M = \{c_k \in \{\text{background, rotation, translation} \}\}$;  
(2) instance masks $I=\{i_k\}^L_{k=1}$ of interactable parts and their motion type label $t_k \in \{\text{ background, rotation, translation} \}\}$;
(3) motion parameters per movable part $m_k$, including motion origin $\{o_k \in \mathbb{R}^3\}$ and axis $\{a_k \in \mathbb{R}^3\}$.
Beyond these tasks, we can also predict connectivity graphs as a separate task based on \protect\OurDataset. This can be achieved using spatial proximity calculations or learning-based methods~(Sec. ~\ref{suppsec:downstream}).

\subsection{Network Architecture}
The overall architecture is shown in Fig.~\ref{fig:usdnet}.
\protect\OurMethod{} extends a Mask3D~\cite{mask3d} backbone with a dense-prediction mechanism to leverage point features for part segmentation and articulation prediction.
\par\noindent\textbf{Part Segmentation and Motion Type Prediction}. 
Following the implementaion in Mask3D, 3D sparse convolution~\cite{minkowski} is applied to extract the point-wise features.
Given the point features and initial queries, stacked transformer decoder layers are applied to generate masks of both movable and interactable instances and the instance queries.
The motion types are predicted with MLP layers on instance queries.
Considering small movable and interactable parts (e.g. switches and buttons) in \protect\OurDataset~, we apply the coarse-to-fine learning strategy from \cite{delitzas2024scenefun3d} during training.  
Noticeably, we propose an auxiliary task in which the spatial vector pointing to the part center is predicted for each point of the interactable part, which can speed up the convergence and improve segmentation accuracy.
% "Point-wise center branch" refers to the  Point-wise axis/origin branch means dense prediction of origin and axis of the articulation parameters with point features.
\par\noindent\textbf{Motion Parameter Prediction.}
Given the fact that the motion parameter is closely related to the geometry and semantic and spatial attributes of the corresponding movable part, we integrate query-based prediction with point-wise dense prediction to better capture those attributes for parameter prediction.
As illustrated by the red blocks in Fig.~\ref{fig:usdnet}, the masked point features of the movable parts are passed through the "Point-wise Axis and Origin Branch" formed by MLPs to produce point-wise predictions of the axis and origins motion parameters.  These predictions are then averaged, with the mean value used for subsequent processing.
On the other hand, movable part queries also go through the MLPs~(yellow block in the figure) to produce an axis prediction. This prediction is averaged with the point features-based  mean prediction.
By doing so, both local features~(point features) and global context~(queries) are leveraged for the motion parameter prediction.
\par
% In general, the key differences of \protect\OurMethod{} from the method proposed in ~\cite{delitzas2024scenefun3d} are that:
% 1) \protect\OurMethod{} simultaneously predicts movable and interactable parts and motion paramter prediction; 
% and 2) we introduce a point-wise dense prediction mechanism~(instead of a query-based prediction) which is illustrated below.
\protect\OurMethod{} differs from the method proposed in ~\cite{delitzas2024scenefun3d} in two key aspects: 1) \protect\OurMethod{} simultaneously predicts movable and interactable parts and motion paramter; and 2) we introduce a point-wise dense prediction mechanism~(instead of a query-based prediction),  which we illustrate below.

\subsection{Implementation Details}
\par\noindent\textbf{Losses.}
For part segmentation, we apply dice loss and binary cross-entropy loss for mask generation: $L_{seg} = \lambda_{dice}L_{dice} + \lambda_{ce}L_{ce}$.
Additional grouping loss is applied for auxiliary task of interactable part grouping: $L_{aux} = \lambda_{aux}\cdot  \sum_{p\in \mathbf{i_k}}  | \mathbf{v_p^*} - \mathbf{v_p} | $, where $\mathbf{v_p^*}$ and $\mathbf{v_p}$ are ground truth and predicted point-to-part-center vector of point $p$ belonging to interactable part $i_k$.
For motion type prediction, we apply cross-entropy loss $\lambda_{cls}L_{cls}$.
For motion parameter prediction, we apply category-specific articulation loss.
For translation motion type, we consider the axis loss, formulated as the cos-dissimilarity between the predicted axis and the ground truth: $L_{arti} = \lambda_{arti} (1 - \cos{<\mathbf{a_k}, \mathbf{a^*_k}>})$.
For rotation motion type, we also calculate the distance of the predicted origin to the ground truth axis:
$L_{arti} = \lambda_{arti} (1 - \cos{<\mathbf{a_k}, \mathbf{a^*_k}>}) + \lambda_{arti} \|\mathbf{a^*_k} \times (\mathbf{o_k} - \mathbf{o^*_k}) \| $.

% \par\noindent\textbf{Parameters.}
\par\noindent\textbf{Training.}
We weight losses during training $L = L_{seg} + L_{cls} + L_{aux} + L_{arti}$ with 
$\lambda_{dice}=2.0, \lambda_{ce}=5.0, \lambda_{cls}=2.0, \lambda_{aux}=1.0, \lambda_{arti}=1.0$.
\protect\OurMethod~ is firstly trained for 1160 epochs on part segmentation tasks and then further trained on joint prediction of part segmentation and motion parameters for 680 epochs. 
The training batch size is 1, on 1 NVIDIA A100-40g GPU and the learning rate is 0.0001.
The input point cloud is cropped with cuboid of $6\times6\;m^2$ to save memory during training. 

% \par
% \todo{specify the input and out}
% \todo{write the descriptions of each module/branch in Fig.3}
% \subsection{Losses}
% \subsection{}
\section{Experiments}\label{sec:experiment}

% \todo{A brief description of the tasks we consider in the evaluation}
We compare \protect\OurMethod~ against baselines on \protect\OurDataset. We also evaluate \protect\OurMethod~ on existing datasets~\cite{delitzas2024scenefun3d, mao2022multiscan} on the task of part segmentation and motion parameter prediction.
More implementation details can be found in Sec. \ref{suppsec:experiment}.
\subsection{Baselines}
As representative approaches for the task of semantic-instasnce segmentation, Softgroup~\cite{vu2022softgroup} and Mask3D~\cite{mask3d}~ are adapted as baselines in our evaluation.
We denote these adapted baselines as SoftGroup$^\dagger$ and Mask3D$^\dagger$. 
For fair comparison, losses of the baselines are the same as \protect\OurMethod{}.
\par\noindent\textbf{SoftGroup$^\dagger$} re-uses the network of SoftGroup for instance segmentation of movable and interactable parts.
Origin and axis prediction branches are implemented to aggregate per-part backbone features for motion parameter prediction.
\par\noindent\textbf{Mask3D$^\dagger$}. 
We apply Mask3D for movable and interactable part segmentation with its original 3D instance segmentation framework.
For articulation parameter prediction, inspired by OPDMulti~\cite{opdmulti} and SceneFun3D~\cite{delitzas2024scenefun3d}, we add an additional head that maps the per-part instance queries to motion parameter predictions. 
% Similar to SoftGroup$^\dagger$, an separate Mask3D-based framework is applied to implement interaction part segmentation.
% \todo{move this part to each subsection if they differ across different tasks}

\subsection{Movable Part Segmentation}
\par\noindent\textbf{Evaluation Metric}.
To evaluate the 3D part segmentation and motion type classification accuracy, we report the mean average precision~(AP) and AP with IoU thresholds of 0.25~($AP_{25}$) and 0.5~($AP_{50}$).
\par\noindent\textbf{Results}.
The performance of the benchmarked methods, reported in Tab. \ref{table:MovSegOurs}, is the highest for \protect\OurMethod~both for $AP_{50}$ and $AP_{25}$.
SoftGroup$^\dagger$ shows the lowest $AP_{50}$ and $AP_{25}$ but the highest $AP$, due to the IoU prediction branch in \cite{vu2022softgroup}, which filters out false positive instance proposals. 
In contrast, Mask3D-based methods are prone to over-confident predictions, as instance confidence is directly estimated from category and mask scores.

\begin{table}[tb]
  \centering
  \small
  % \begin{tabular}{l  cc  cc}
  %    \multirow{2}{*}{Method} & \multicolumn{2}{c}{$AP_{50}$} & \multicolumn{2}{c}{$AP_{25}$}  \\
  %      & rot & trans & rot & trans  \\
  \begin{tabular}{l c c c}
  \toprule
     Method & $AP$ & $AP_{50}$ &$AP_{25}$  \\
       % & rot & trans & rot & trans  \\
    \midrule
    SoftGroup$^\dagger$ & \textbf{22.7} & 32.7 & 37.2\\
    Mask3D$^\dagger$~(baseline) & 18.1 & 39.1 & 58.9 \\
    \protect\OurMethod~ (ours) &  19.8 & \textbf{41.8} & \textbf{59.9} \\
    \bottomrule
  \end{tabular}
      \vspace{-8pt}
    \caption{Segmentation of movable part on \protect\OurDataset.} \label{table:MovSegOurs}
    \vspace{-10pt}
\end{table}

\subsection{Interactable Part Segmentation}
\par\noindent\textbf{Evaluation Metric}.
Similarly, we report $AP_{25}$ and $AP_{50}$ for measuring accuracy of interactable part segmentation.
\par\noindent\textbf{Results}.
Tab. \ref{table:InteractionSegOurs} shows that \protect\OurMethod~performs better than Mask3D$^\dagger$, illustrating the effectiveness of the auxiliary task of the point-center-shift prediction as shown in Fig.~\ref{fig:usdnet}. 
It is noticeable that SoftGroup$^\dagger$ performs inferior to \protect\OurMethod~and Mask3D$^\dagger$, which is in accordance with the benchmarks of \cite{dai2017scannet, yeshwanthliu2023scannetpp, delitzas2024scenefun3d}.

\begin{table}[tb]
  \centering
  \small
  \begin{tabular}{l  c  c c}
  \toprule
     Method & $AP$ & $AP_{50}$ & $AP_{25}$\\
    \midrule
    SoftGroup$^\dagger$ & 6.8 & 14.5 & 25.4 \\
    Mask3D$^\dagger$~(baseline) &  \textbf{12.7} & 30.2 & 55.6 \\
    \protect\OurMethod~(ours) &  \textbf{12.7} & \textbf{31.1} & \textbf{55.9} \\
    \bottomrule
  \end{tabular}
        \vspace{-8pt}
    \caption{Segmentation of interactable part on \protect\OurDataset.} \label{table:InteractionSegOurs}
    \vspace{-8pt}
\end{table}

\subsection{Motion Parameter Prediction }
\par\noindent\textbf{Evaluation Metric}.
Inspired by \cite{delitzas2024scenefun3d}, we extend $AP_{50}$ with our own articulation loss to evaluate the accuracy of motion parameter prediction given the part segmentation prediction. 
As shown in Table~\ref{table:MovArtiOurs}, $AP_{50} + \{\text{Axis, Origin}\}$ represents that a prediction is considered as true positive only when (1) the IoU between part segmentation and G.T. mask is greater than $50\%$, 
% and (2) the articulation loss is below certain threshold. \TODO{define the threshold or provide more information}
Following ~\cite{delitzas2024scenefun3d}, the threshold for axis is $1-cos15^\circ$ and for origin is $0.25 m$.
\par\noindent\textbf{Results}.
Table~\ref{table:MovArtiOurs} shows that \protect\OurMethod{} achieves significantly better performance than the baselines in prediction of both motion parameters by a significant margin~(surpassing SoftGroup$^\dagger$ by 7.3\% and Mask3D$^\dagger$ by 5.7\% for $AP_{50}+\text{Axis+Origin}$).
We further evaluate \protect\OurMethod{} on the MultiScan and SceneFun3D datasets. As shown in Tab. \ref{table:evalotherdatasets},  \protect\OurMethod{} outperforms the baselines.
As illustrated in later ablation study and supplementary material, its advantage lies in the the dense articulation prediction that better integrate geometrical and spatial information. 

\begin{table}[tb]
  \centering
  \small
  \begin{tabular}{l  ccc ccc}
  \toprule
     \multirow{2}{*}{Method} & \multicolumn{3}{c}{$AP_{50}$} \\
     & +Origin & +Axis & +Origin+Axis   \\
     
    \midrule
    SoftGroup$^\dagger$ & 18.5 & 21.5 & 17.7 \\
    Mask3D$^\dagger$~(baseline)  & 24.4 & 33.8 & 19.3 \\
    \protect\OurMethod~(ours)  & \textbf{31.4} & \textbf{34.6} & \textbf{25.0} \\
    \bottomrule
  \end{tabular}
        \vspace{-8pt}
    \caption{Motion parameters of movable part on \protect\OurDataset.} \label{table:MovArtiOurs}
    \vspace{-12pt}
\end{table}

\subsection{Ablation Study}
To evaluate the impact of dense point-wise predictions in \protect\OurMethod, we performed ablation studies (Tab. \ref{table:ablationsdense}). Results demonstrate: (1) without the dense prediction of axis articulation, both origin and axis prediction performance drop; (2) without the dense prediction of origin, the axis prediction benefits but the origin prediction is significantly affected,  leading to a substantial decline in joint articulation prediction; (3)  combining both dense prediction mechanisms yields optimal joint articulation accuracy.
\begin{table}[tb]
  \centering
  \small
  \begin{tabular}{l  ccc ccc}
  \toprule
     \multirow{2}{*}{Method} & \multicolumn{3}{c}{$AP_{50}$} \\
      & +Origin & +Axis & +Origin+Axis   \\
       \midrule
       Mask3D$^\dagger$~(wo. both)  & 24.4 & 33.8 & 19.3 \\
      wo. dense axis & 26.9 & 30.3  & 21.9 \\
     wo. dense origin & 21.1 & \textbf{38.7} & 18.2 \\
     \protect\OurMethod~  & \textbf{31.4} & 34.6 & \textbf{25.0} \\
    \bottomrule
  \end{tabular}
        \vspace{-8pt}
    \caption{Ablations on dense articulation prediction mechanisim.} 
          \vspace{-8pt}\label{table:ablationsdense}
\end{table}

\begin{table}[tb]
  \centering
  \small
    \begin{tabular}{c  c c c}
        \toprule
        Dataset & SoftGroup$^{\dagger}$ & Mask3D$^\dagger$ &\protect\OurMethod{}.  \\
        \midrule
        % \hline
        MultiScan & 4.7 & 23.3 & \textbf{26.0} \\
        SceneFun3D & 12.8 & 22.4 & \textbf{30.5}\\
       \bottomrule
        % \hline
    \end{tabular}
\vspace{-8pt}
\caption{$AP_{50}+Origin+Axis$ on Multiscan and SceneFun3D}
\label{table:evalotherdatasets}
\vspace{-16pt}
\end{table}

\section{Downstream Tasks}\label{sec:applicationmainpaper}
% By providing the 3D USD-based scenes with rich annotations, \protect\OurDataset{} enables rich downstream applications.
The rich 3D annotations in USD format provided by  \protect\OurDataset~ unlock a range of novel applications, such as cross-dataset and cross-domain generalization, scene editing, and robotic manipulation, as demonstrated here.

\begin{table}[tb]
  \centering
  \small
    \begin{tabular}{c  c c c c}
        \toprule
        Settings & AP50 & +Origin & +Axis & +Origin+Axis.  \\
        \midrule
        % \hline
        wo. pretrain & 34.8 & 28.1 & 30.1 & 24.3 \\
        w. pretrain & \textbf{40.5} & \textbf{31.3} & \textbf{33.8} & \textbf{26.0}\\
       \bottomrule
        % \hline
    \end{tabular}
\vspace{-8pt}
\caption{
\textbf{Pre-training on \protect\OurDataset{} improves performance on scene understanding task.} 
Evaluation on articulation prediction on Multiscan~\cite{mao2022multiscan} with / without pre-training on \protect\OurDataset. 
}
\label{table:pretrianmultiscan}
\vspace{-8pt}
\end{table}

% \subsection{Pretraining}
% \par\noindent\textbf{Improved Performance through Pre-training.}
\par\noindent\textbf{Cross-Dataset Generalization.}
To demonstrate \protect\OurDataset's generalization capabilities for scene understanding, we evaluated \protect\OurMethod~ on movable part segmentation and motion parameter prediction on the MultiScan~\cite{mao2022multiscan} dataset.
Pre-training on \protect\OurDataset significantly improved performance on both tasks (Tab. \ref{table:pretrianmultiscan}). 
% \todo{which method is used here}
% can be used in pre-training for downstream 3D scene understanding tasks.
% In  we can see that by pretraining on \protect\OurDataset, we can significantly improve the performance of \protect\OurMethod~on the benchmarks in .
% \noindent\textbf{Part Seg\&Articulation Pred}:
% To show the benefits of \protect\OurDataset~for the tasks in other datasets, we pretrain \protect\OurMethod~on \protect\OurDataset~and finetune it on MultiScan for the movable part segmentation and articulation parameter prediction. 

\par\noindent\textbf{Cross-Domain Generalization.}
We illustrate the value of \protect\OurDataset{} in addressing the sim-real gap using URDFormer~\cite{chen2024urdformer} as a case study.
URDFormer~\cite{chen2024urdformer} predicts URDFs from a single image by
% is a framework for 2D-to-URDF prediction with a single image, 
training solely on auto-generated synthetic data. 
URDF (Unified Robot Description Format) defines a scene description format for simulators, however, its support for complex meshes and collision geometries is less comprehensive than that of USD.
% We conduct experiments on \protect\OurDataset{} and Multiscan datasets and demonstrate that fine-tuning URDFormer on \protect\OurDataset~ can significant improve by a large margin its performance on URDF prediction of real-world, as shown in Table~
Specifically, we evaluate URDFormer on the part segmentation task using both \protect\OurDataset~ and MultiScan. 
%
% We evaluate part-detection accuracy of URDFormer on \protect\OurDataset~ and MultiScan datasets, comparing generated URDFs of articulated objects against g.t. annotations. 
The achieved 22.7 $AP_{50}$ on MultiScan and 16.4  $AP_{50}$  on the more diverse \protect\OurDataset{} (Tab. \ref{tab:urdformer}) highlight challenges with real-world data. 
Finetuning on \protect\OurDataset{} leads to notable improvements,
% improves the $AP_{50}$ score on MultiScan to 35.1, 
demonstrating the value of \protect\OurDataset{} for enhancing real-world adaptability. More details are provided in the Sup. Mat.
% \begingroup
% \setlength{\intextsep}{1pt}
\begin{table}[tb]
    \centering
    {\small % Apply tiny inside the scope
    % \renewcommand{\arraystretch}{0.8} % Adjust row height
    % \resizebox{\columnwidth}{!}{%
    \begin{tabular}{ c  c c }
        \hline
        Eval Dataset & Zero-Shot & Finetuned on Articulate3D  \\
        %\hline
        \midrule
        Multiscan & 22.7 & \textbf{35.1} \textcolor{ForestGreen}{(+12.4)} \\
        \protect\OurDataset{} & 16.4 & \textbf{38.2} \textcolor{ForestGreen}{(+21.8)} \\
        \hline
    \end{tabular}
    % }
    }
\vspace{-8pt}
\caption{\textbf{Pre-training on \protect\OurDataset{} improves performance of method trained only on synthetic data.} Part detection accuracy~($AP_{50} (\%)$) of URDFormer prediction on real-world datasets.
Improvement reported in green.}
\label{tab:urdformer}
\vspace{-16pt}
\end{table}

% \subsection{LLM-based Scene Editing}
\par
\begin{figure}
  \centering
  \begin{subfigure}[b]{0.15\textwidth}
    \centering
    \includegraphics[width=\linewidth]{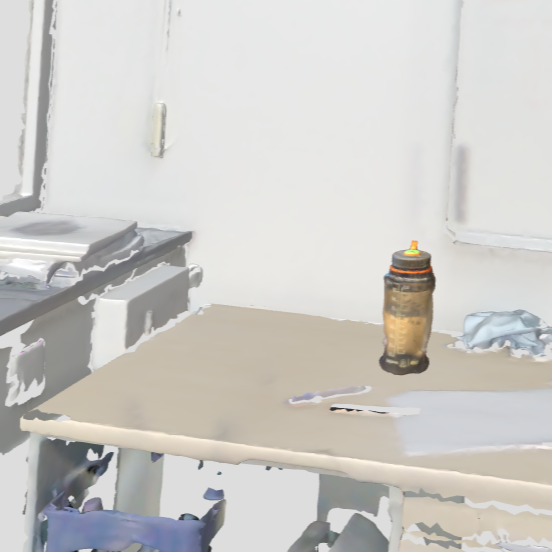}
    \caption{Bottle Insertion}
    \label{fig:subfig1}
  \end{subfigure}
  \hfill
  \begin{subfigure}[b]{0.15\textwidth}
    \centering
    \includegraphics[width=\linewidth]{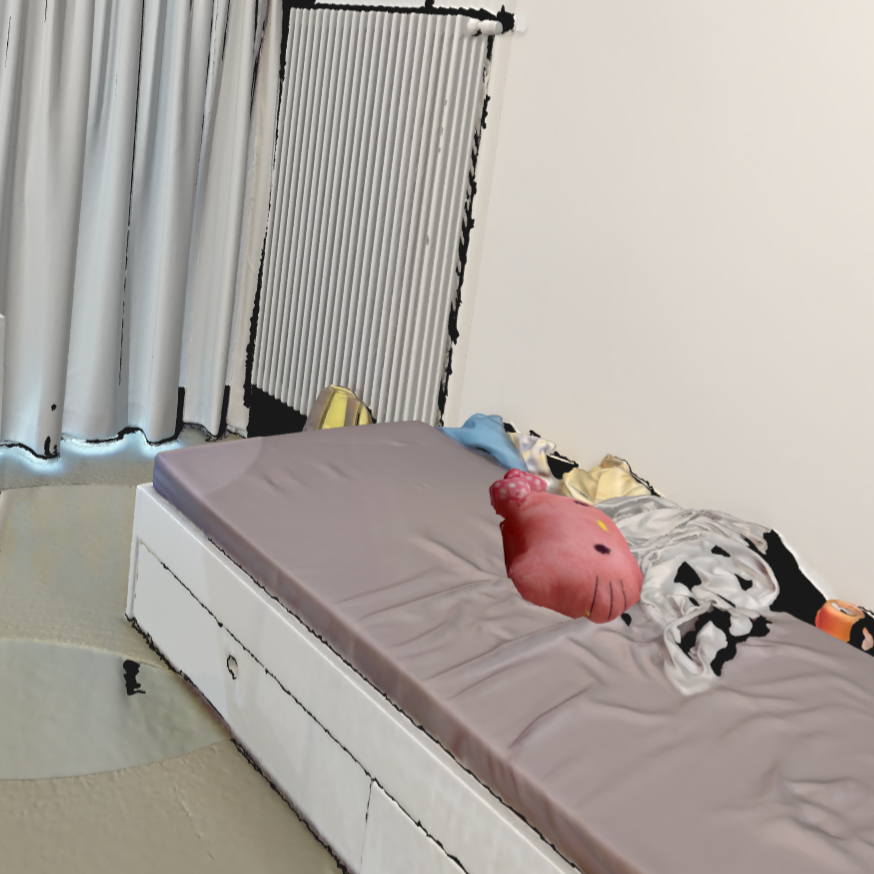}
    \caption{Pillow Insertion}
    \label{fig:subfig2}
  \end{subfigure}
  \hfill
  \begin{subfigure}[b]{0.15\textwidth}
    \centering
    \includegraphics[width=\linewidth]{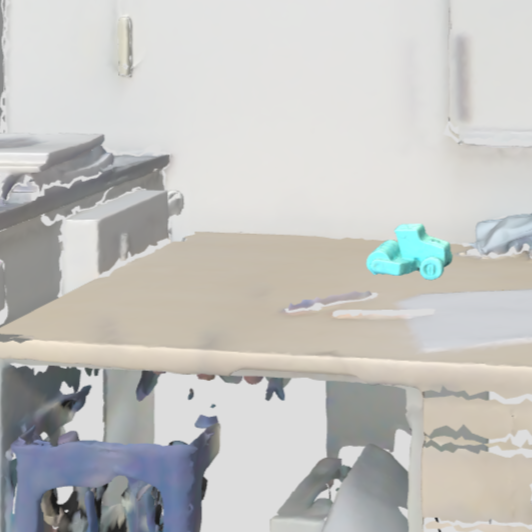}
    \caption{Toy car Insertion}
    \label{fig:subfig3}
  \end{subfigure}
  \vspace{-5pt}
  \caption{\textbf{USD scene editing through LLM prompting.} Visualizations of \protect\OurDataset scenes after object insertion via LLM.}
  \label{fig:usd}
  \vspace{-10pt}
\end{figure}

\noindent\textbf{LLM-based Scene Editing.}
Due to the hierarchical and organized representation of USD, \protect\OurDataset{} scenes can be easily understood by LLMs and edited by users through prompts. 
In Fig. \ref{fig:usd}, we demonstrate that LLMs can be used to perform semantically-aware object insertions in an \protect\OurDataset{} scene. For example, upon prompting an LLM to place a pillow in an appropriate location within a given scene, the LLM generates code whose execution produces the scene with the pillow on the scene's bed. Further details are offered in Sec. \ref{suppsec:sceneinsertion}.
%
% \subsection{Simulation-to-go for Robotics}

\par
\begin{figure}
  \centering
  \begin{subfigure}[b]{0.23\textwidth}
    \centering
    \includegraphics[width=\linewidth]{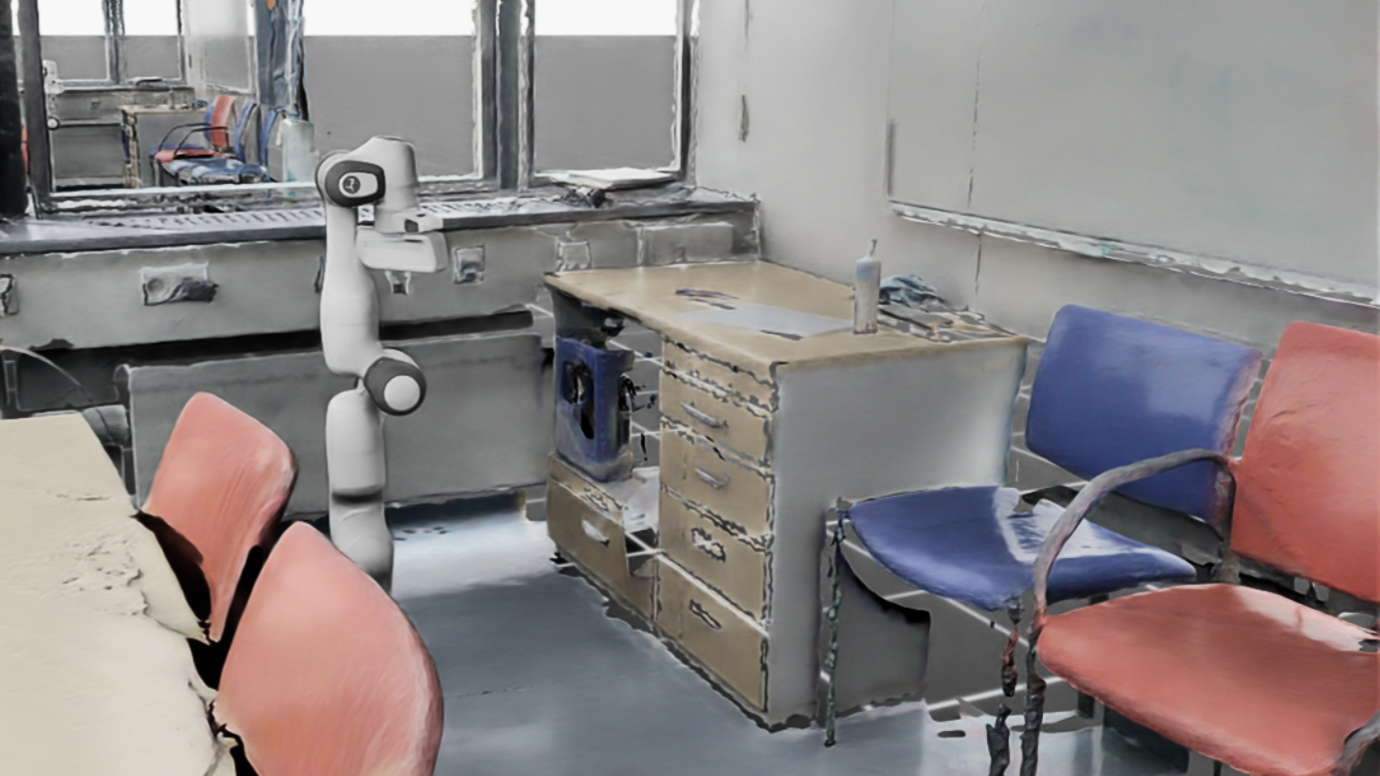}
  \end{subfigure}
  \hfill
  \begin{subfigure}[b]{0.23\textwidth}
    \centering
    \includegraphics[width=\linewidth]{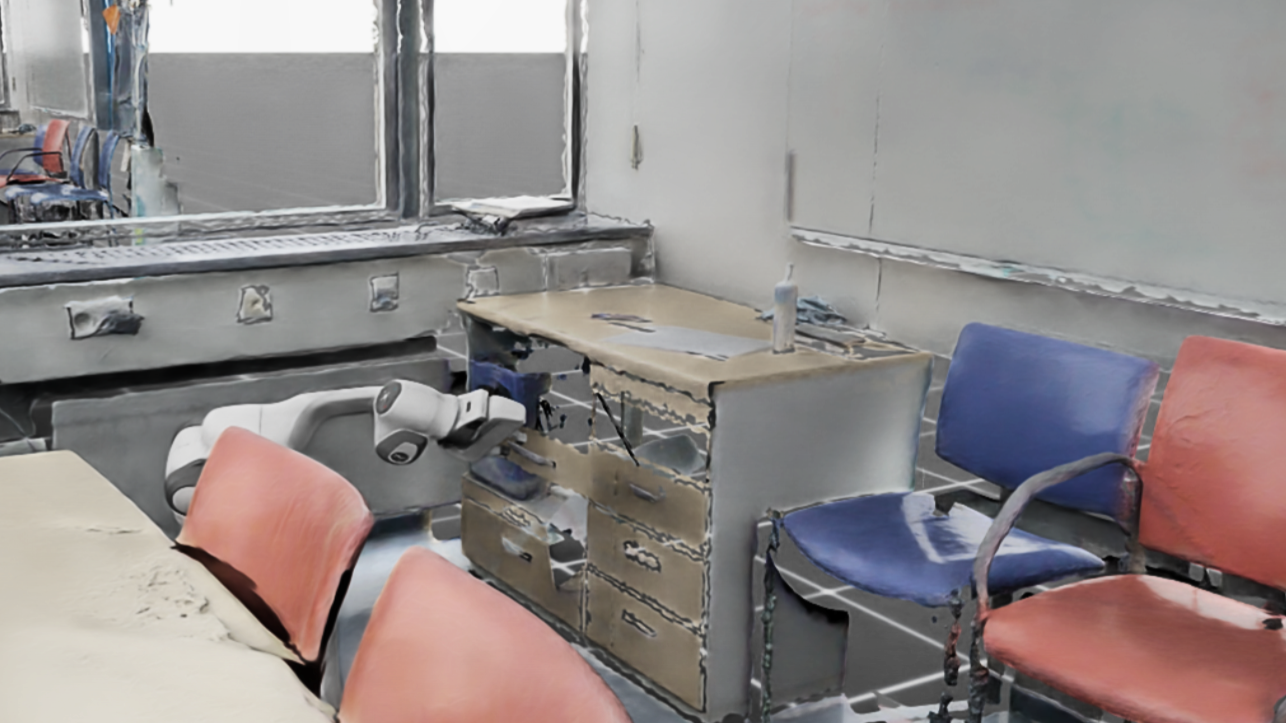}
  \end{subfigure}
  \vspace{-5pt}
  \caption{\textbf{\protect\OurDataset{} enables easy policy training for robotic manipulation in simulation environments.} Drawer opening using a PPO-learned policy: start (left) and end state (right) in an \protect\OurDataset~ scene simulated with IsaacSim and IsaacLab.}
  \label{fig:policy}
  \vspace{-10pt}
\end{figure}

\noindent\textbf{Simulation-to-go for Robotics.}
With the simulation-ready scenes from \protect\OurDataset{}, it is easy to apply robotic policy training. Using only the USD scene representation, \protect\OurDataset{} scenes can be directly uploaded in IsaacSim and simulated for policy learning using IsaacLab \cite{mittal2023orbit}. 
As an example, we train a policy to open a drawer, using PPO \cite{ppo2017Schulman}, as shown in Fig. \ref{fig:policy}. We can also easily apply planner-based policies. We note that \protect\OurDataset{} is currently the only real-world scene-level dataset that includes the full specter of annotations necessary for direct usage in simulation.  For more details, we refer to Sec. \ref{sec:policy}.
%

%
% \subsection{Addressing Sim-Real Gap}

%\endgroup 

\section{Conclusion}
We present \protect\OurDataset, a dataset of 280 high-quality scans of real-world indoor scenes, annotated with semantic segmentations at the object and part levels, as well as articulation, connectivity, and object mass information. \protect\OurDataset~ provides the most comprehensive set of articulation annotation categories to date,  through specification of  motion parameters, movable, interactable and fixed parts. \protect\OurDataset~ is the first large-scale, non-synthetic dataset supporting physical scene simulation, while also offering greater variability than synthetic alternatives. The practical value of \protect\OurDataset~ is demonstrated through its application in various downstream tasks, including LLM-based scene editing, physical simulation for embodied AI, and reducing the sim-to-real gap for articulated asset generation.

To demonstrate \protect\OurDataset's quality and utility, we propose \protect\OurMethod, a novel method that jointly predicts 
3D movable and interactable part segmentation and articulation from 3D scene point clouds. It introduces a dense point-wise prediction strategy for improved articulation parameter estimation.  Further evaluations on diverse datasets validate \protect\OurMethod's effectiveness.
\section*{Acknowledgments}
This research was partially funded by the Ministry of Education and Science of Bulgaria (support for INSAIT, part of the Bulgarian National Roadmap for Research Infrastructure). This project was supported with computational resources provided by Google Cloud Platform (GCP).
% {
%     \small
%     \bibliographystyle{ieeenat_fullname}
%     \bibliography{main}
% }

{
    \small
    \bibliographystyle{ieeenat_fullname}
    \bibliography{main}
}

\clearpage
\maketitlesupplementary

\noindent
We provide additional details on: 
\begin{itemize}[left=2em]
    \item the annotation process~(Sec.~\ref{suppsec:annotationprocess}
),
    \item the statistics of \protect\OurDataset~(Sec.~\ref{suppsec:datasetstatistics}),
    \item applications of \protect\OurDataset~(Sec.~\ref{suppsec:downstream}),
    % \item connectivity graph prediction~(Sec.~\ref{suppsec:connectivitypredict}),
    \item experiment details and additional results~(Sec.~\ref{suppsec:experiment}).   
\end{itemize}

\section{Annotation Process}
\label{suppsec:annotationprocess}
\subsection{Annotation Tool}
\begin{figure}
  \centering
  \begin{subfigure}[b]{0.45\textwidth}
    \centering
    \includegraphics[width=\linewidth]{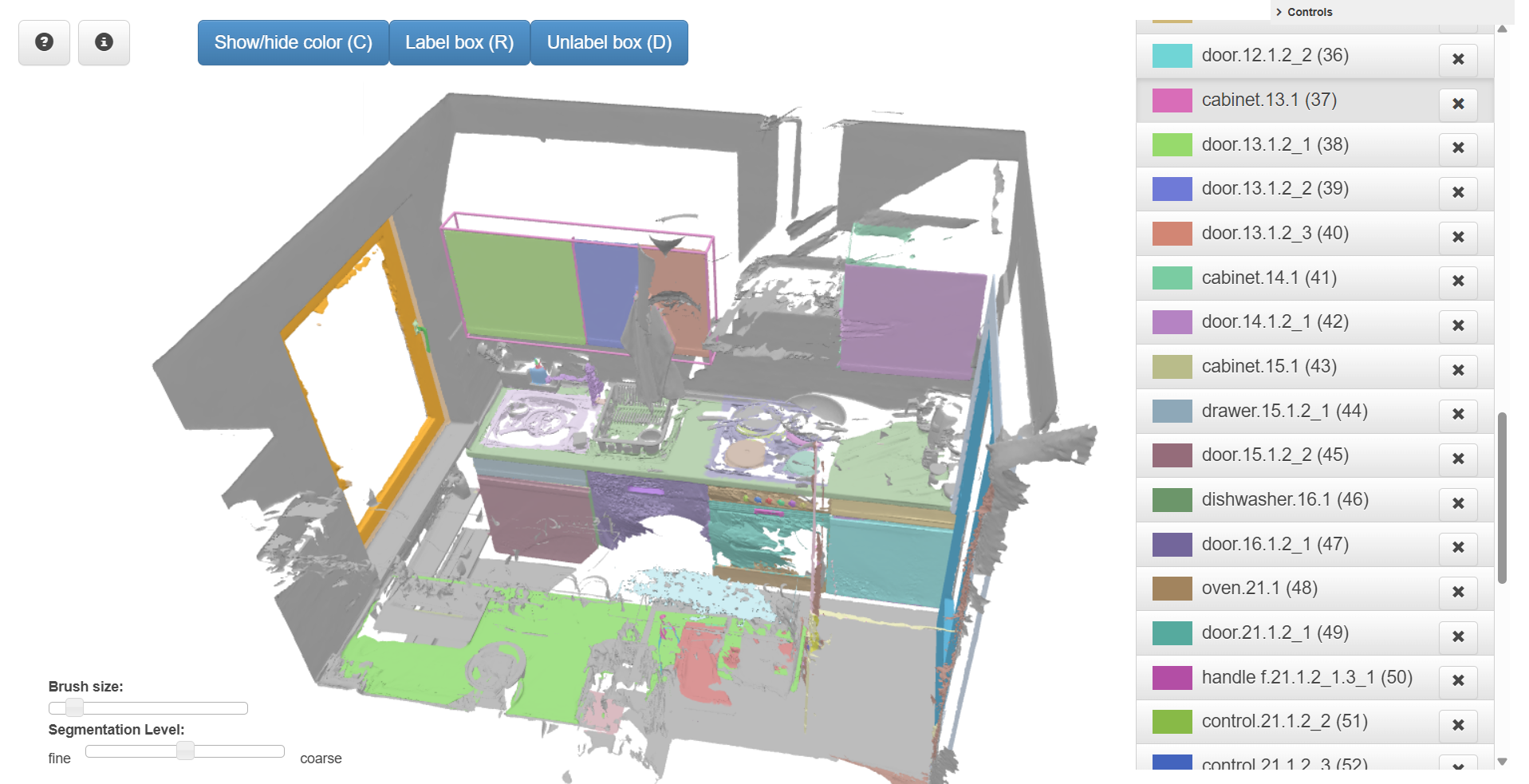}
    \caption{Segmentation interface for annotation of object and part semantic segmentation and connectivity.}
  \end{subfigure}
  \begin{subfigure}[b]{0.45\textwidth}
    \centering
    \includegraphics[width=\linewidth]{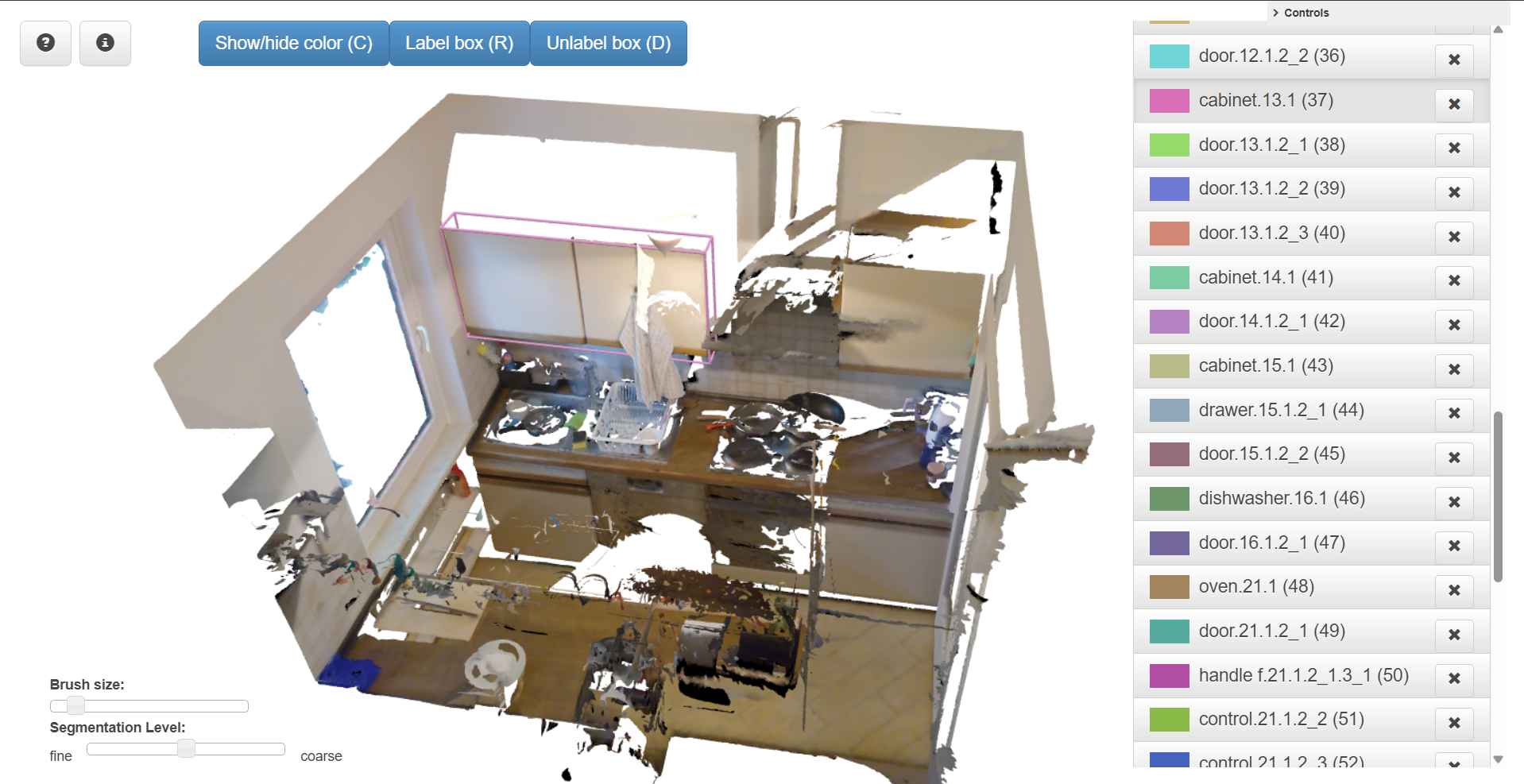}
    \caption{The segmentation interface offers switching between color and segmentation view, enabling distinction of finer details.}
  \end{subfigure}
  \begin{subfigure}[b]{0.45\textwidth}
    \centering
    \includegraphics[width=\linewidth]{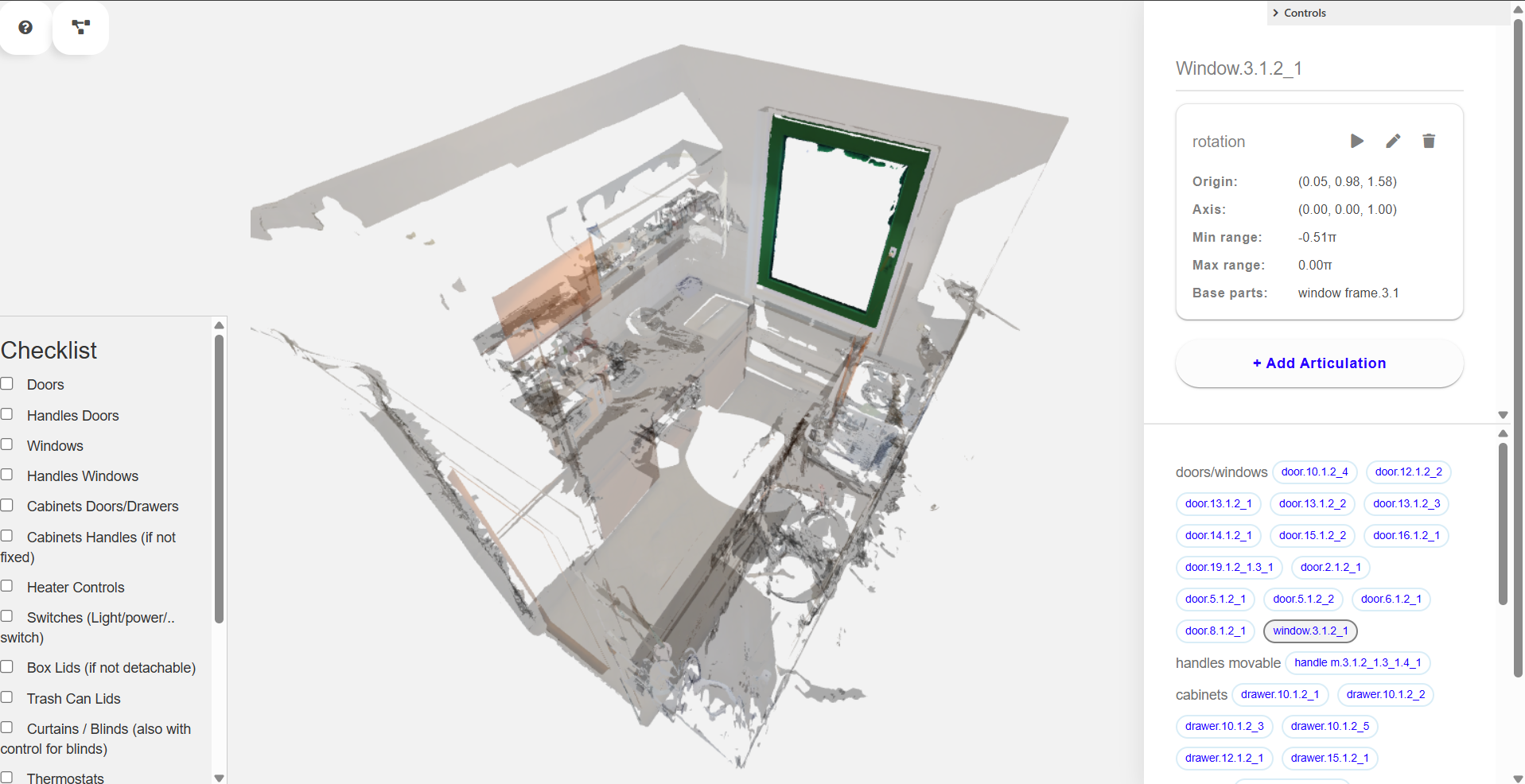}
    \caption{Articulation annotation interface. The annotators are provided with a checklist of part categories to annotate, and a sorted by label list of segmented parts, promoting a systematic annotation process.}
  \end{subfigure}
  \begin{subfigure}[b]{0.45\textwidth}
    \centering
    \includegraphics[width=\linewidth]{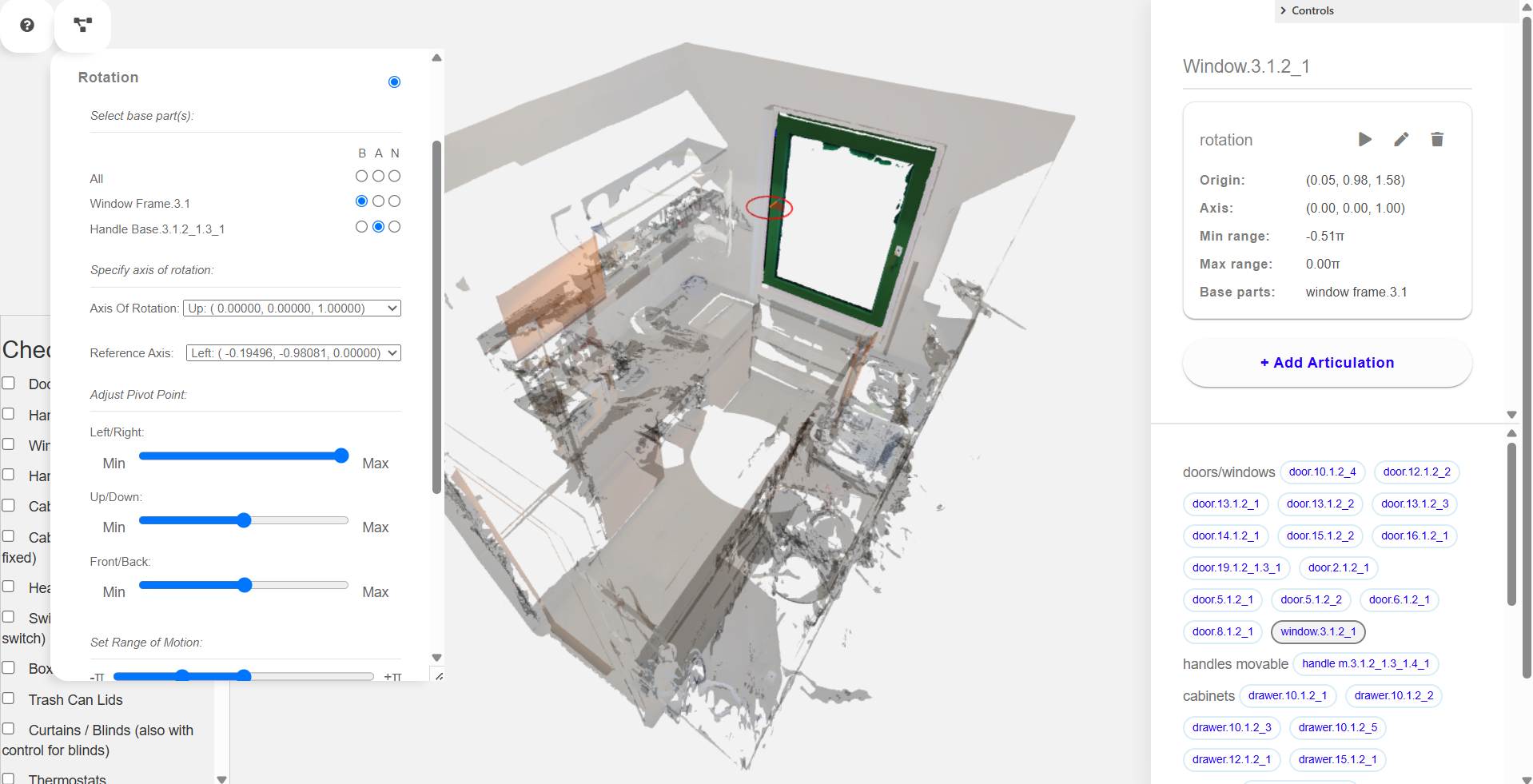}
    \caption{Annotation of motion parameters.}
  \end{subfigure}
  \vspace{-10pt}
  \caption{The annotation tool.}
  \label{fig:tool}
\end{figure}

We introduce the interface of the annotation tool used to create \protect\OurDataset~ in Fig. \ref{fig:tool}. It features instance semantic segmentation functionalities, extended to enable connectivity annotation, as well as a view for articulation annotations.

\begin{figure}
  \centering
  
  \begin{subfigure}[b]{0.22\textwidth}
    \centering
    \includegraphics[width=\linewidth]{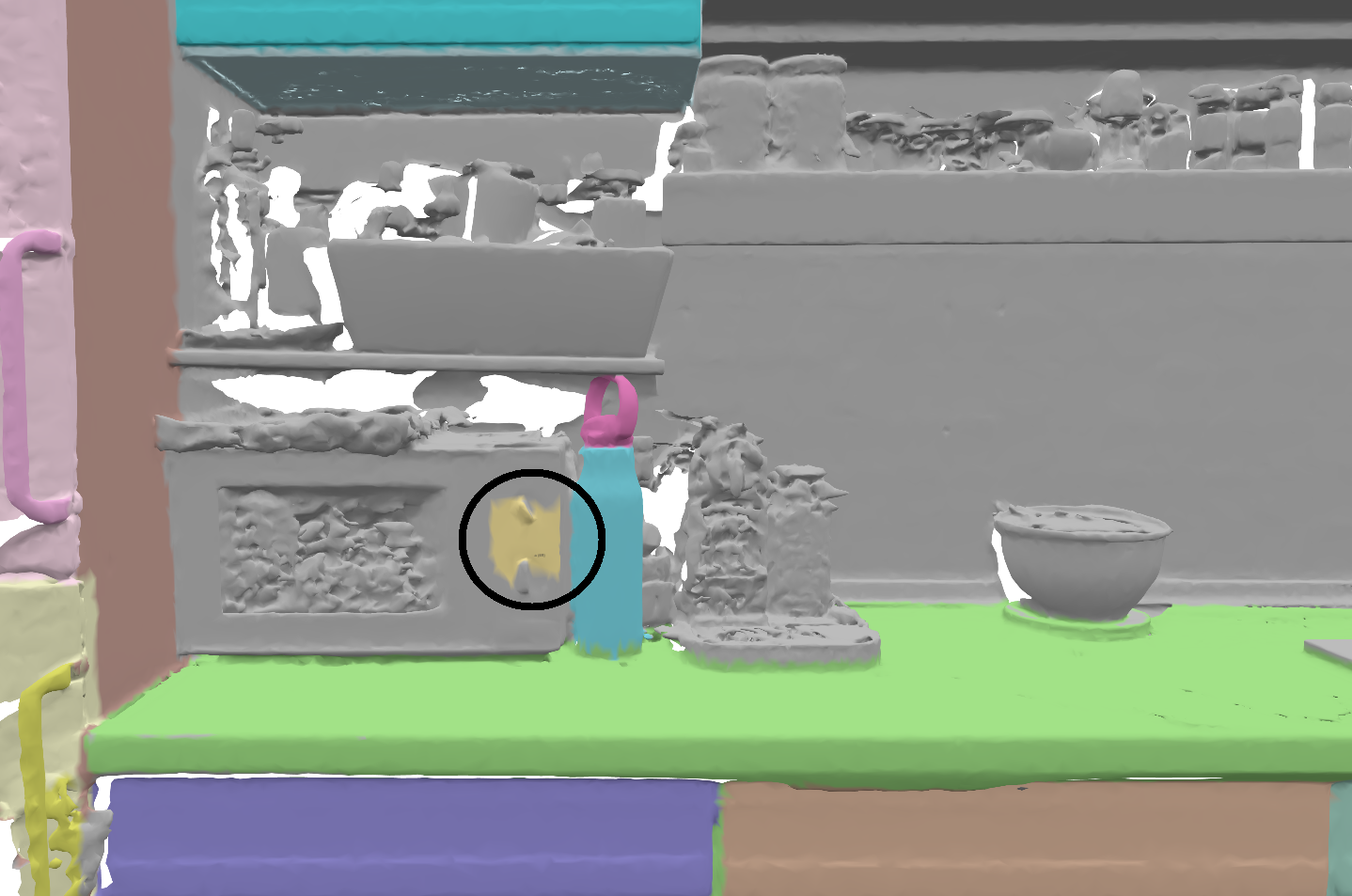}
  \end{subfigure}
  \hfill
  \begin{subfigure}[b]{0.22\textwidth}
    \centering
    \includegraphics[width=\linewidth]{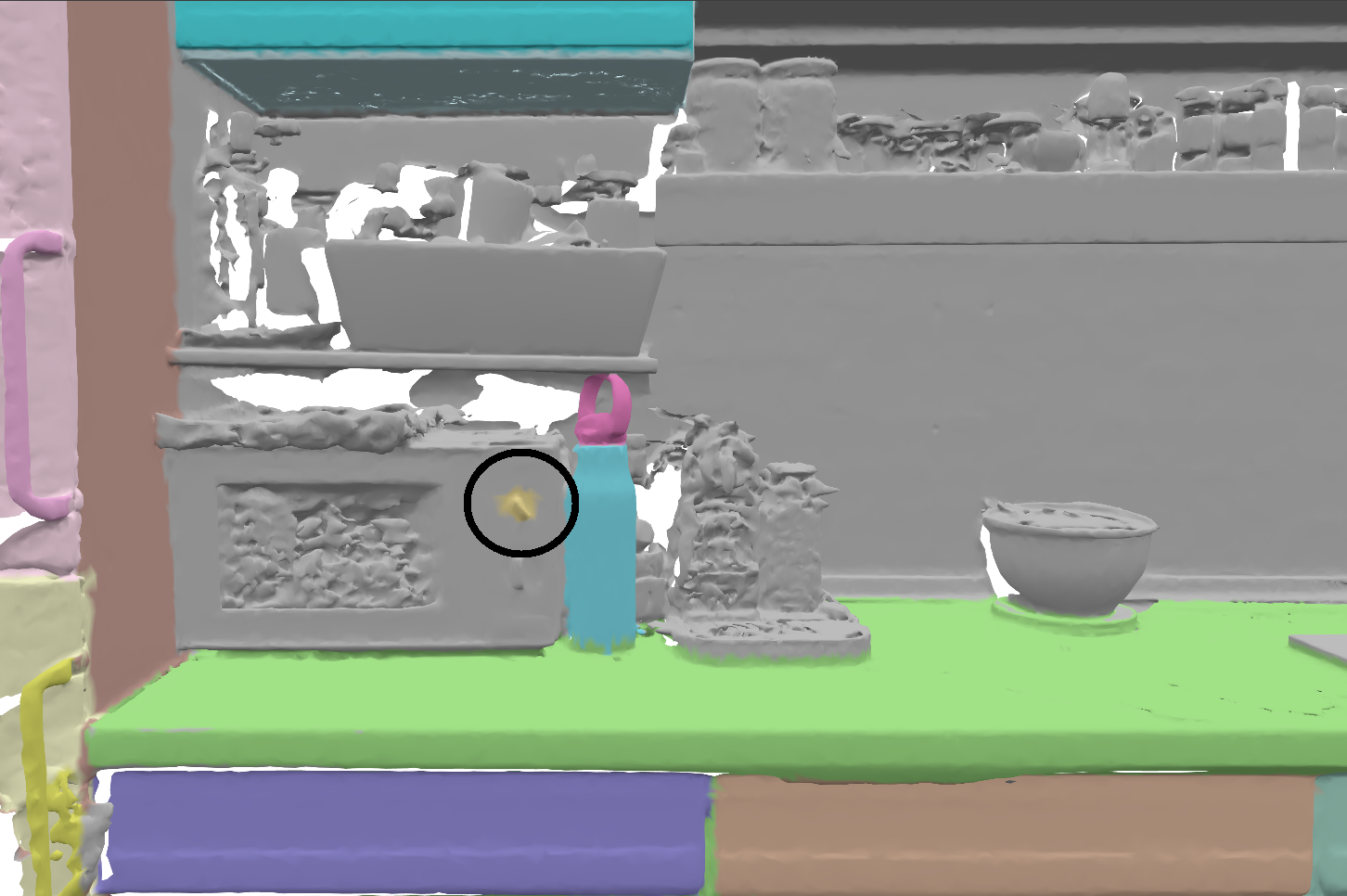}
  \end{subfigure}
  \vspace{-10pt}
  \caption{Segmentation result of smaller details using the provided oversegmentation (left) vs annotating on fine-grained level (right). Segments produced by the oversegmentation can prove unable to capture fine variations in the geometry, showing the need for a fine-grained segmentation support.}
  \label{fig:overseg}
\end{figure}

An essential feature for the annotation of \protect\OurDataset~ was the support both for fine-grained segmentation, as well as for an oversegmentation-based annotation. While the provided initial segments support fast and accurate annotation for big objects with flat surfaces, e.g. cabinets, fine-grained details such as buttons, knobs, switches, etc., are not recognized, as seen in Fig. \ref{fig:overseg}.

\subsection{Annotators Training}\label{suppsec:annotatortraning}
To ensure the quality and consistency of annotations, annotators underwent comprehensive training supported by detailed documentation and tutorials. The preparation materials included:
\par
\noindent\textbf{Segmentation Training:}
\begin{itemize}
    \item Two training videos, totaling 40 minutes, covering the segmentation tool, illustrative examples of typical cases and exceptions.
    \item Documentation with examples for all object classes in the fixed object label list, including: (1) images of segmented parts, (2) their corresponding connectivity graphs, and (3) the step-by-step sequence of commands used to achieve precise segmentation.
    \item A complete guide to the application's controls and shortcuts.
\end{itemize}

\par
\noindent\textbf{Articulation Training:}
\begin{itemize}
    \item A 15-minute training video demonstrating the articulation annotation tool and process, with examples of rotation and translation articulations.
    \item Documentation with examples of the different articulated parts classes.
\end{itemize}

Additionally, annotators were provided with chat support throughout the annotation process. The chat support was especially valuable during the initial review phase, allowing annotators to resolve questions, particularly around complex connectivity graphs, e.g., oven models, where controls are physically attached to the oven door but semantically part of the oven body.

% \subsection{Annotation Process}
% The annotation process was structured into two stages:

% \par
% \noindent\textbf{Segmentation and Connectivity Labeling:} 
% Initial annotations were performed by the primary annotators.
% A dedicated sixth annotator conducted quality control, ensuring adherence to guidelines.
% For minor corrections, the sixth annotator directly refined the annotations. For more substantial issues, the data was returned to the original annotator for revision.
% Upon the sixth annotator’s approval, the segmentation and connectivity annotations were finalized and submitted for the next stage.

% \par
% \noindent\textbf{Articulation Annotation:}
% Only after the segmentation and connectivity labeling was approved, the data was submitted for articulation annotation.
% The sixth annotator was again tasked to approve or reject submissions for this stage upon a detailed examination.

\section{Dataset Statistics}\label{suppsec:datasetstatistics}
\begin{figure*}
    \centering
    \includegraphics[width=\linewidth]{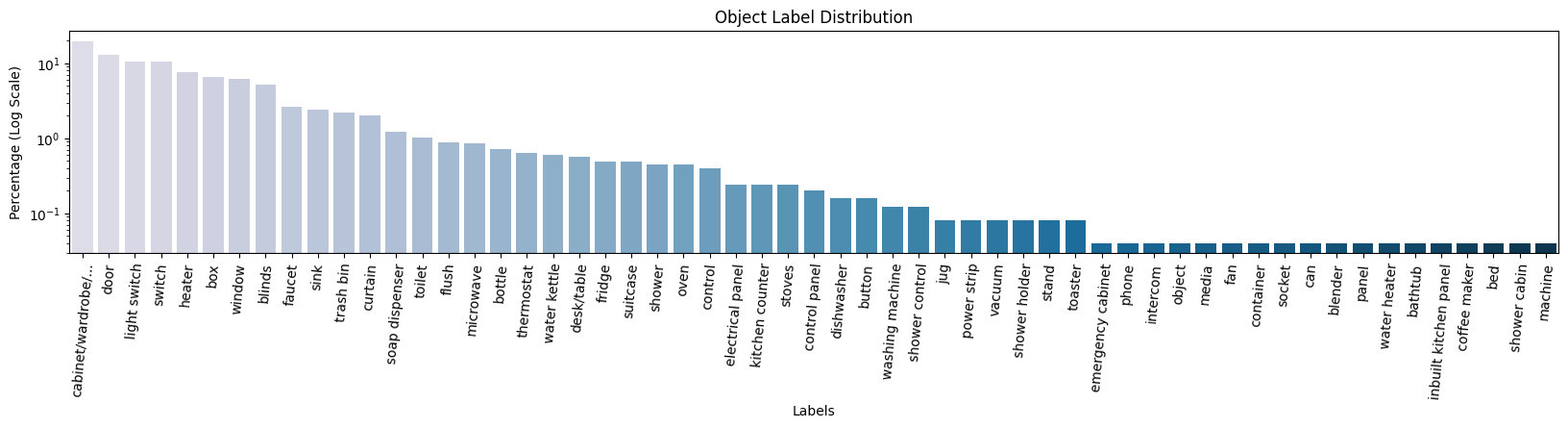}
     \vspace{-18pt}
    \caption{Distribution of the object-level labels.}
    \vspace{-5pt}
    \label{fig:objdist}
\end{figure*}

\begin{figure*}
    \centering
    \includegraphics[width=\linewidth]{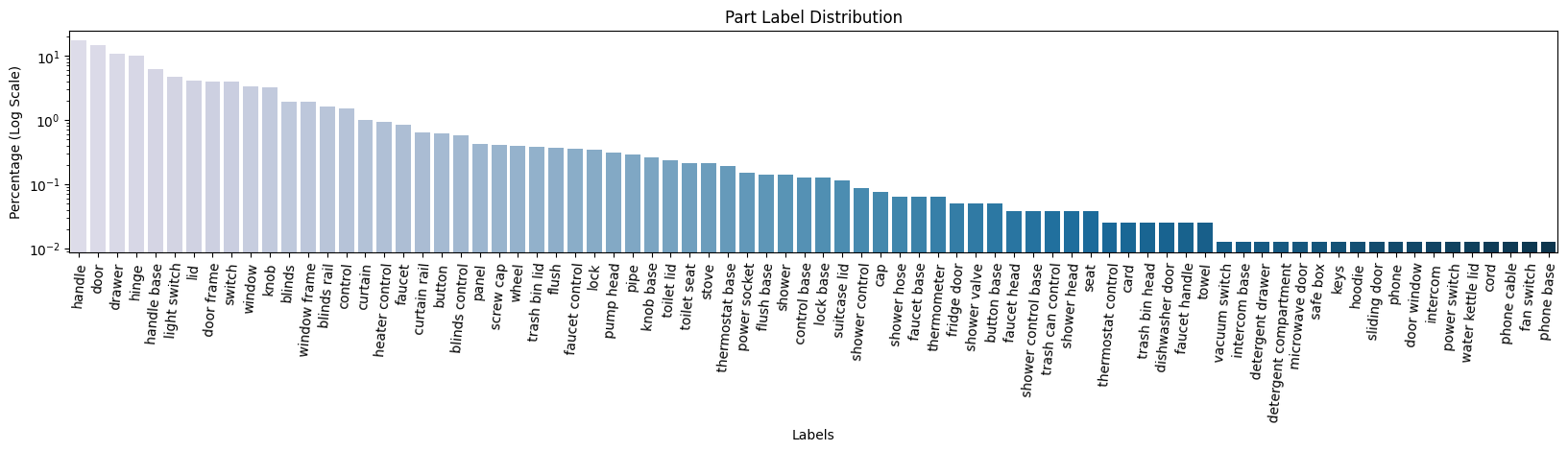}
     \vspace{-18pt}
    \caption{Distribution of the part-level labels.}
    \vspace{-8pt}
    \label{fig:partdist}
\end{figure*}

\begin{figure}
  \centering
  \begin{subfigure}[b]{0.45\textwidth}
    \centering
    \includegraphics[width=\linewidth]{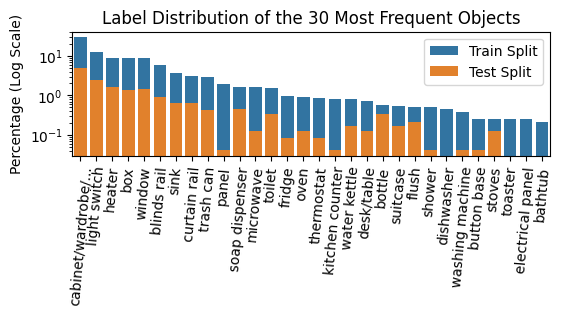}
  \end{subfigure}
  \hfill
  \begin{subfigure}[b]{0.45\textwidth}
    \centering
    \includegraphics[width=\linewidth]{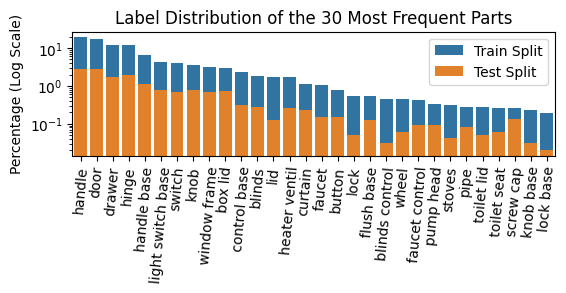}
  \end{subfigure}
  \vspace{-10pt}
  \caption{Distributions of the 30 most frequent object and part labels, per split.}
  \label{fig:statslabels}
\end{figure}

For a comprehensive overview of the annotated object and part labels, we refer to Fig. \ref{fig:objdist} and Fig. \ref{fig:partdist}, which provide information both on the labels and on their distributions. Distributions of the 30 most frequent labels per train and test split are shown in Fig. \ref{fig:statslabels}. 

We note that the dataset also includes additional labels to address variations based on location (e.g., "cabinet" versus "wardrobe"), functional mechanisms (e.g., "faucet handle" versus "faucet ventil"), and other contextual distinctions which are not captured in the provided figures. We will include resources for mapping these additional labels and their parent categories with the dataset release.

\begin{figure*}
    \centering
    \includegraphics[width=\linewidth]{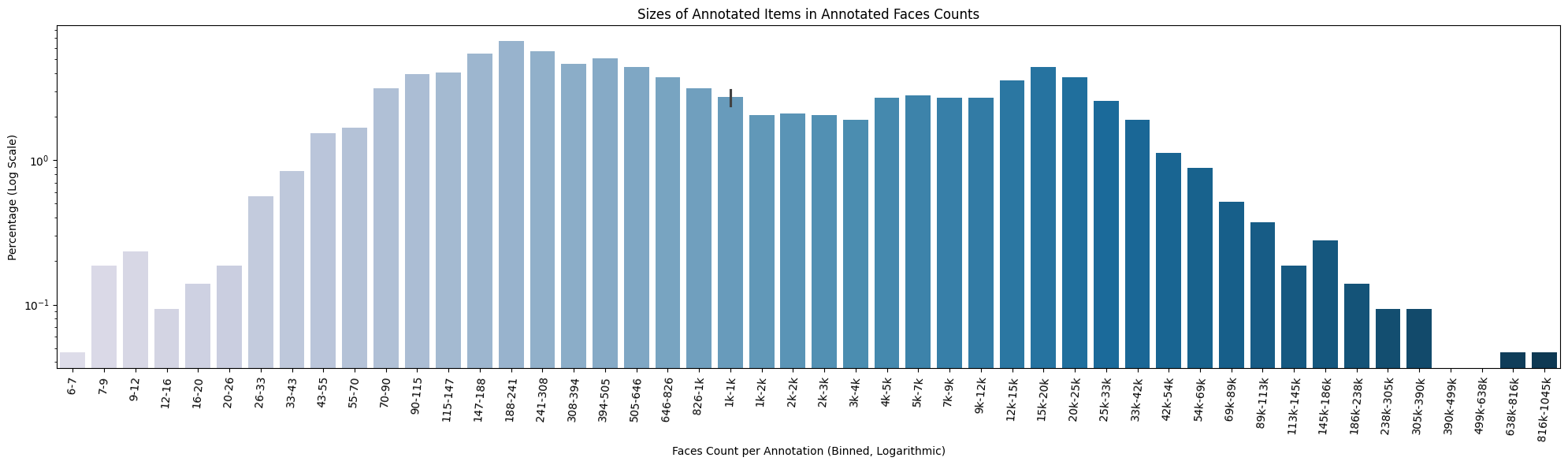}
     \vspace{-15pt}
    \caption{Distribution of face counts per annotated item using logarithmic binning. \protect\OurDataset~ features numerous high-detail small annotations alongside larger objects}
    \label{fig:sizesfaces}
\end{figure*}

We also provide size information for the annotated items (both objects and parts) in Fig. \ref{fig:sizesfaces}, where size is measured by the faces (triangles) count of each item. Our analysis reveals that \protect\OurDataset~ includes a mix of large objects and many smaller ones with only a few faces. This highlights both the high quality of the annotations in capturing intricate details, as well as the variability within the dataset.

\begin{figure*}
    \centering
    \includegraphics[width=\linewidth]{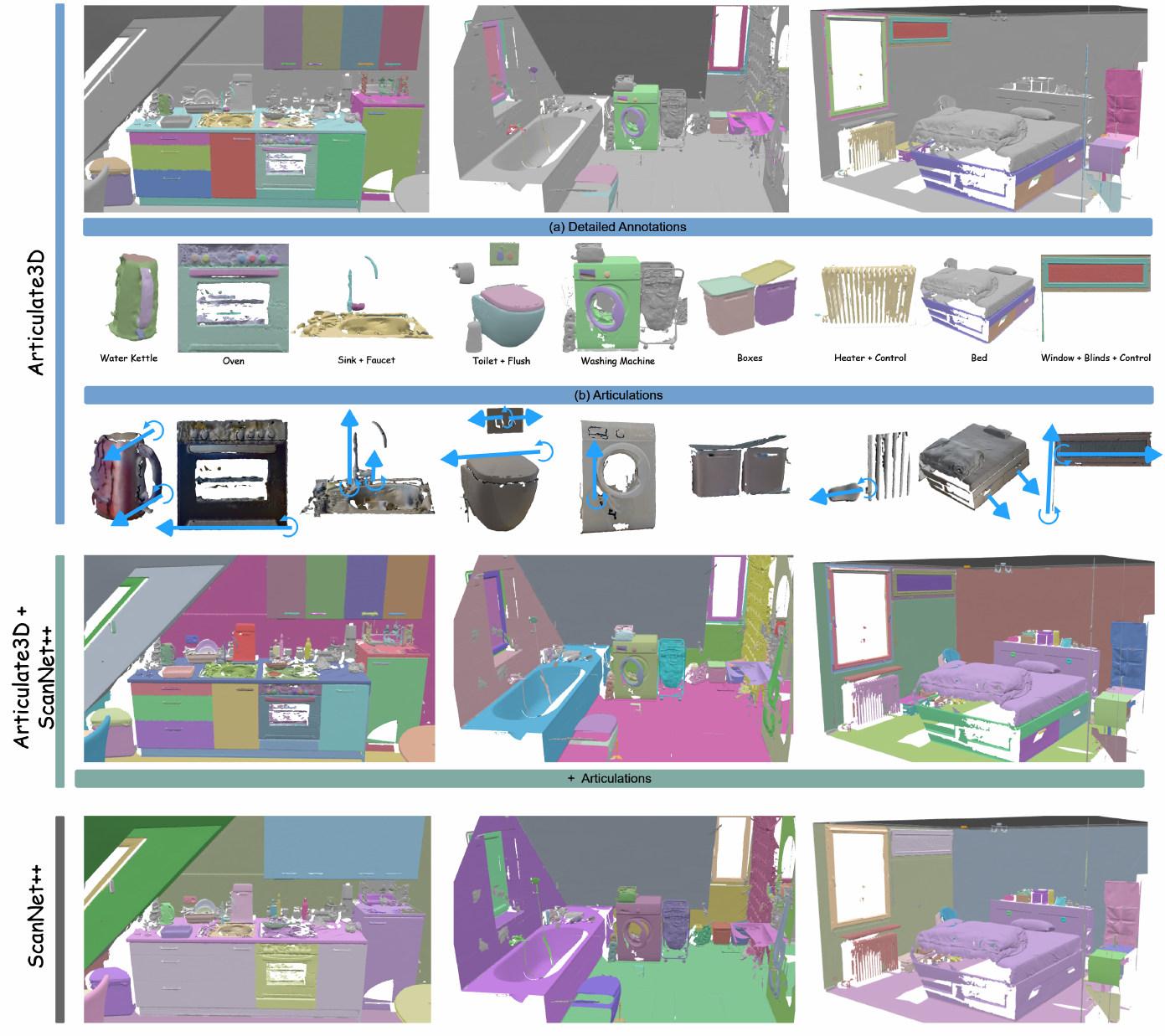}
    \caption{Overview of the annotations provided by \protect\OurDataset, ScanNet++ \cite{yeshwanthliu2023scannetpp}, and their combination. \protect\OurDataset~ offers (a) detailed annotations at both the object and part levels, including connectivity information, and (b) full articulation annotations. By combining ScanNet++ with \protect\OurDataset, users gain a comprehensive scene segmentation with nearly 100\% coverage, alongside detailed annotations for interactable objects and their motion specifications.}
    \label{fig:examples}
\end{figure*}

Our statistical analysis focuses solely on annotations added by our annotators. Since our work builds upon the ScanNet++ dataset \cite{yeshwanthliu2023scannetpp}, we ensure full compatibility between our annotations and those from ScanNet++. Unaltered ScanNet++ annotations (e.g., walls, floors, and sofas) are directly integrated into the final USD, while certain labels, such as "cabinet," are entirely replaced by our own annotations, as illustrated in Fig. \ref{fig:examples}. We will provide a script merging the annotations from \protect\OurDataset~ and ScanNet++.

\subsection{Segmentations Quality Evaluation}
To assess the quality of our segmentation annotations, we conducted a control experiment involving two annotators. They were each tasked with re-annotating a "control scene"—a scene that had been previously annotated, then reviewed, and approved by the sixth annotator. The annotations were manually matched item-by-item between the two annotators, and IoU was calculated for each matched pair. The scene's average IoU served as the metric for annotation quality. 
% Temporary draft, TODO determined
Notably, the re-annotators were distinct from the original annotators.
We achieved an IoU of 0.93 for the control scene.
%
% The first control scene was selected randomly. Upon observing that its scan poorly captured small details such as handles, we selected a second scene with better-captured small objects. This selection involved random sampling until a scene meeting the criteria was found. Importantly, we ensured that the annotators performing the re-annotations were not the original annotators for these scenes.

% For the scene with a high-quality scan, we achieved an IoU of 0.93. In contrast, the challenging scene yielded an IoU of 0.81. A closer examination of item-wise IoU values for the challenging scene revealed that the lower score was exclusively due to small objects with poor mesh quality. For example, handles with sparse and irregular face distributions significantly impacted the IoU, as even minor variations in the annotated face sets caused substantial changes in the metric.

\section{Downstream Applications}
\label{suppsec:downstream}
\subsection{LLM-based Scene Editing}
\label{suppsec:sceneinsertion}
We present a pipeline for automated, semantically-aware object insertion into USD scenes. Given a USD scene from \protect\OurDataset, a 3D object file, and the object's label, our solution produces a new USD scene with the object placed in a semantically-appropriate location. For example, in a bedroom scene, a pillow object would be placed on a bed, while in an office scene, a bottle would be placed on a desk. The pipeline uses a LLM to show the USD-understanding capabilities of LLMs, as well as to minimize user involvement. Implementation details are outlined below.

The method requires three inputs: (1) a USD \protect\OurDataset~ scene, (2) a 3D object file, and (3) the object’s label. We support various 3D file formats (e.g., OBJ, USD, GLB). Given the inputs, the pipeline extracts item labels and connectivity data from the USD scene and prompts the LLM to determine the appropriate placement target (e.g., a bed for the pillow) and the type of surface required (e.g., horizontal for a pillow, vertical for a poster). With this information, the pipeline employs RANSAC to identify a suitable placement plane on the target object. Using the plane, the input object, and an example USD insertion script defined by us, the LLM generates a script customized for the specific insertion case. The generated script is then executed to produce the updated \protect\OurDataset~ USD scene.

For our pipeline, we have tested two LLMs - GPT-4o mini \cite{gpt40mini} and GPT-4o \cite{gpt4o}, both producing the desired results. We will release the pipeline as a Python CLI library.

\subsection{Simulation-to-go for Robotics}
\label{sec:policy}
Our data can be easily uploaded in a physics simulator and used for robotics policy learning, without manual adaptions. We use IsaacSim with IsaacLab \cite{mittal2023orbit} as simulator setup due to their USD-centricity. The utilized GPU is RTX 3090. All simulations are conducted with the Franka robot.

\begin{figure}
  \centering
  \begin{subfigure}[b]{0.23\textwidth}
    \centering
    \includegraphics[width=\linewidth]{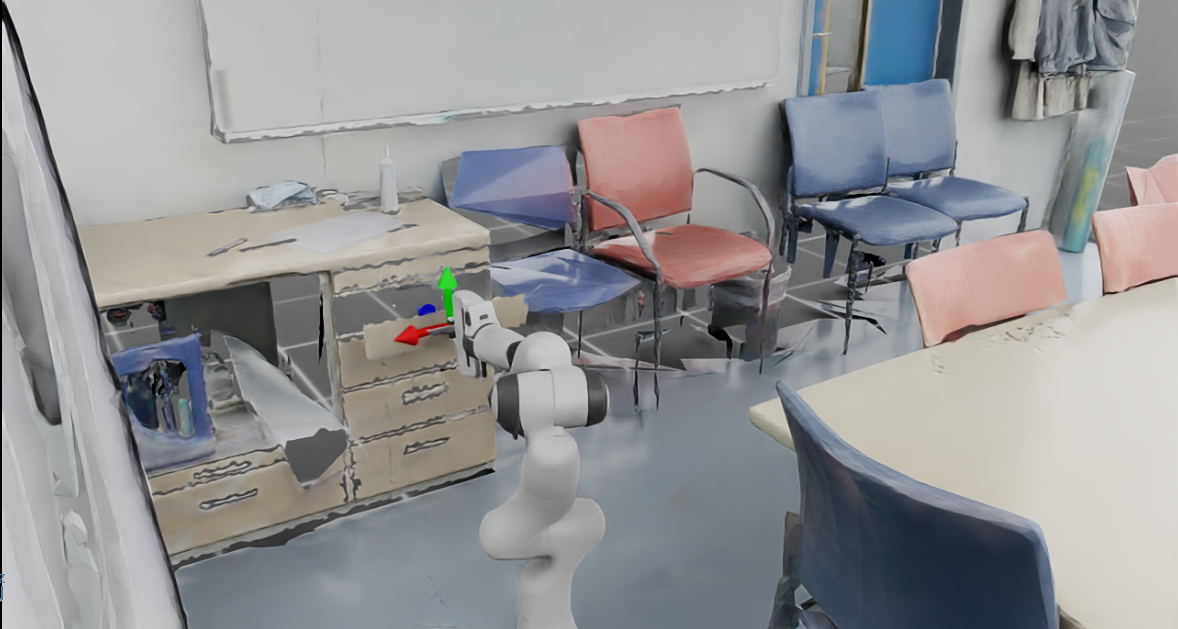}
  \end{subfigure}
  \hfill
  \begin{subfigure}[b]{0.23\textwidth}
    \centering
    \includegraphics[width=\linewidth]{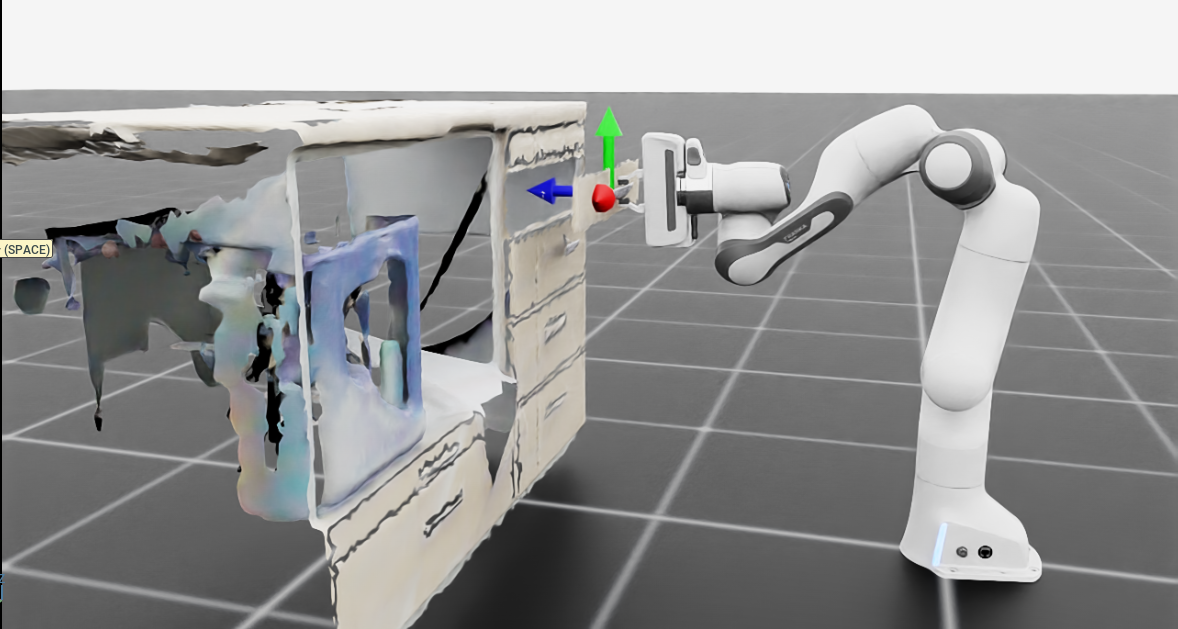}
  \end{subfigure}
  \vspace{-5pt}
  \caption{Simulation of an \protect\OurDataset~ scene (left) vs simulation of a single object from an \protect\OurDataset~ scene (right).}
  
  \label{fig:simobjectvsscene}
  \vspace{-15pt}
\end{figure}

\protect\OurDataset~ scenes are versatile and can be utilized with various policies, as demonstrated by their compatibility with both planner-based solutions and policy training via PPO \cite{ppo2017Schulman}. We also note that the \protect\OurDataset~ scenes can be used both for scene-level simulation, as well as for object-level, as depicted in Fig. \ref{fig:simobjectvsscene}. This is enabled by USD's support for easy extraction of single objects. Scene-level manipulations are also easy as objects within scenes can be removed/added or rearranged to achieve versatility. If using the object extraction strategy, the users obtain more than 3k articulated real-world USD objects from over 50 categories.

\protect\OurDataset~ scenes are decimated using quadratic decimation. In this way we enable learning on multiple environments in parallel, speeding up training. 

\par\noindent\textbf{Planner-based policies.} We apply a planner-based policy only requiring a specification of an object of interest (e.g., a cabinet), a movable part (drawer) and the corresponding interactable part (handle). All articulation parameters and configurations are derived from the USD scene file. 

\par\noindent\textbf{Proximal policy optimization (PPO) policies.} \protect\OurDataset~ scenes support training in simulation, which we demonstrate by learning a drawer-opening policy via PPO, with 87\% success rate. We use 1024 environments and train for 30k iterations with 40 steps per environment, with learning rate 5.0e-4 and adaptive schedule.

\subsection{Cross-Domain Experiment - URDFormer}
As illustrated in Sec.~\ref{sec:applicationmainpaper}, we evaluated URDFormer~\cite{chen2024urdformer} on both \protect\OurDataset{} and Multiscan dataset to demonstrated the value of \protect\OurDataset{} in cross-domain generalization.
URDFormer consist of global scene-object arrangement generation and local object-part generation. 
We focus on the second part and assume the poses of articulated articulated object are given.
To obtain data required for fine-tuning/evaluation of URDFormer, we generate pairs of images and 3D articulated object by selecting 3 top images per object in which the object are most visible and then crop the images to fit the object.
URDFormer introduces biases in the framework design for object URDF generation: it generates bounding boxes of parts of articulated object and then fit the boxes with the part mesh from PartNet-Mobility~\cite{sapien}, and it also assumes the articulation axis is aligned with one of edges of those bounding boxes.
Thus, for fair evaluation, we focused the accuracy of the generated bounding boxes instead of the biased geometry or motion parameter.
Generally, given the cropped image of articulated objects, we apply URDFormer to predict the bounding boxes and motion type~(rotation, translation and background) of the movalble parts and evaluate the part detection accuracy by comparing the predicted part bboxes and the ground truth ones.

\subsection{Connectivity Graph Prediction}\label{suppsec:connectivitypredict}
To demonstrate the value of \protect\OurDataset~ in 3D holistic scene understanding, particularly in the context of connectivity graphs within articulated objects, we conduct experiments on the task of connectivity graph prediction.
As introduced in Sec.~\ref{sec:builddataset}, a connectivity graph refers to the hierarchies between different partegments.
In the graph, the ”root” part is the base to which other parts are attached. 
A ”child” refers to a part that belongs to the parent part.
We train a network on \protect\OurDataset~that takes the 3D point cloud of objects' parts as input and predicts the connectivity graph of the parts by inferring relationship between each part pair. 
% \subsection{Method}
\par
\begin{figure}[t] 
\begin{center}
\includegraphics[width=0.99\columnwidth]{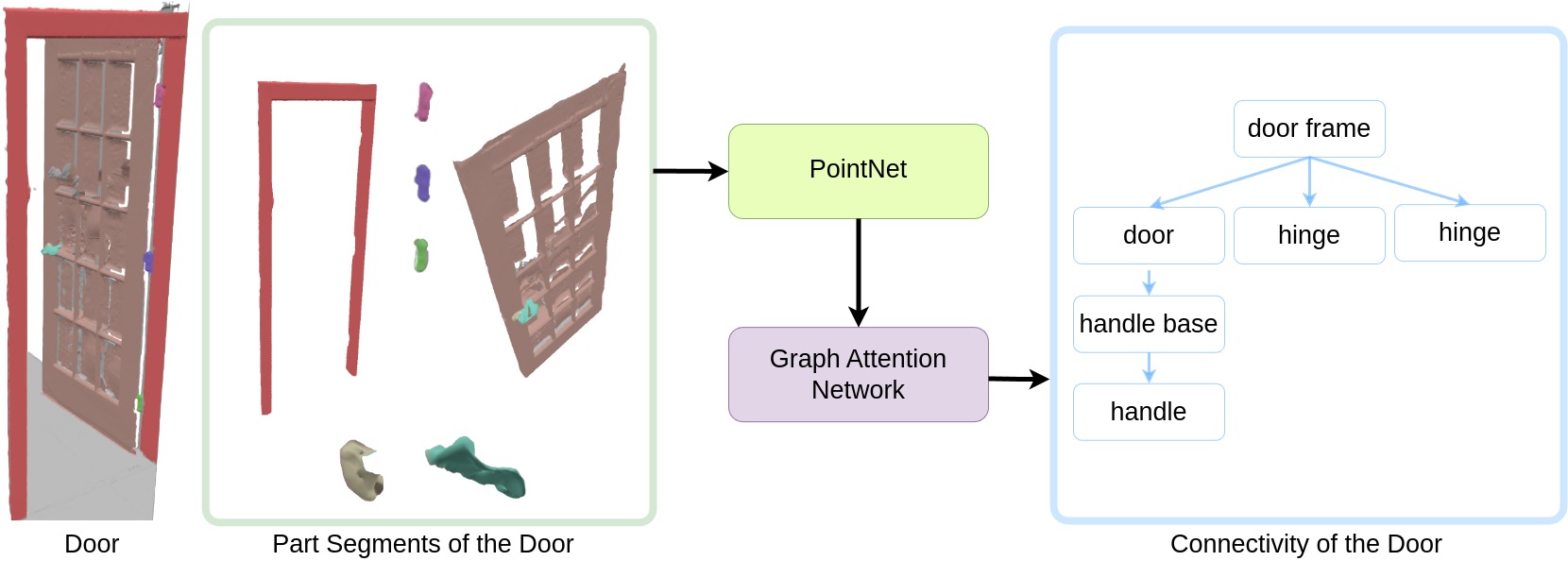}
    \caption{Connectivity Graph Prediction of Parts of Articulated Objects.}\label{fig:connectivityprediction}
\end{center}
\vspace{-20pt}
\end{figure}

\noindent\textbf{Method.}
% We build a simple network that takes in the 3D pointcloud of part segments of objects as input and predict the connectivity graph of the objects. 
% Given the part segments of articulated objects, we implement a simple network to predict the relationships between parts and infer connectivity graphs.
We draw inspiration from the existing scene-graph-based framework~\cite{sarkar2023sgaligner3dscene} and apply a similar network for the task.
As shown in Fig.~\ref{fig:connectivityprediction}, PointNet~\cite{pointnet} is applied to extract the information of geometry and appearance of part segments, outputting per-part feature vectors. 
Then we treat each part segment as a node and feed the feature vectors into 
a graph attention network~\cite{2018graphattentionnetworks} to extract the relationship between the nodes.
The features of the two node are concatenated and fed into a MLP to infer the pairwise relationship between them~($\{no\; relationship,\; parent\; of,\; child\; of\}$).
% \subsection{Result}
\par
\begin{figure}
    \centering
    % First image
    \includegraphics[width=0.85\columnwidth]{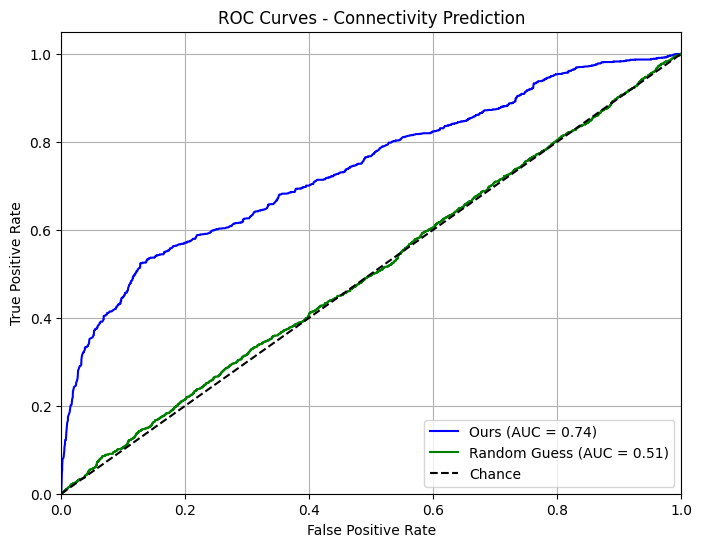}
    \vspace{0.5cm} % Add space between the images
    % Second image
    \includegraphics[width=0.85\columnwidth]{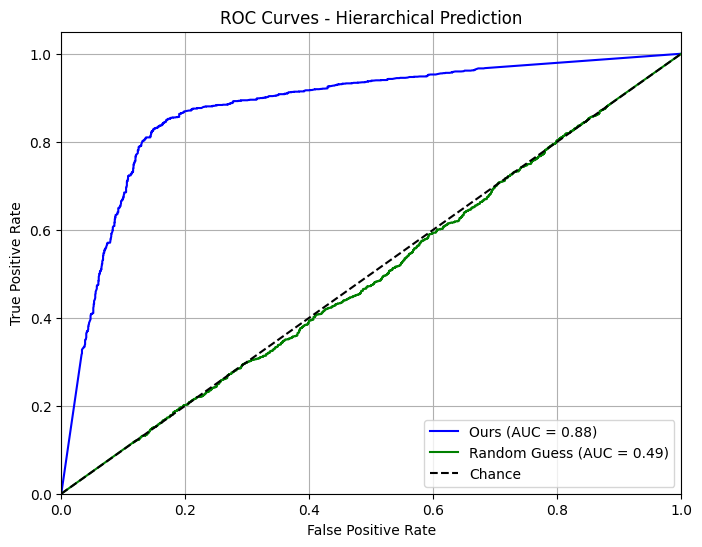}
    \vspace{-20pt}
    \caption{ROC curve of connectivity and hierarchical relationship of part segments prediction of our method.} 
    \label{fig:roccurves}
    \vspace{-8pt}
\end{figure}

\noindent\textbf{Result.}
In Table.~\ref{table:ConnectPred}, we show the accuracy of the connectivity graph prediction. From the table we can see that (1) the random guess could achieve $36\,\%$ accuracy of edge prediction by randomly picking one relationship from $\{no\; relationship,\; parent\; of,\; child\; of\}$ for each pair of part segments, and (2) it has poor performance in the connectivity graph prediction for objects as only $6.1\%$ of objects' connectivity graph are correctly predicted.
Compared to the random guess, by training on \protect\OurDataset~, our network can achieve much better performance in both edge prediction~($72.7\%$) and connectivity graph prediction~($31.1\%$).  
% \par
In order to further understand the performance of our model in connectivity or edge prediction, we plot the ROC curves as shown in Fig.~\ref{fig:roccurves}.
As the edge prediction is a 3-category classification problem, we transform the results into two binary classification problems for straightforward ROC plot: connectivity classification~($\{no\; relationship, \{\; parent\; of,\; child\; of\} \}$) and hierarchical classification~($\{\; parent\; of,\; child\; of\} \}$).
From Fig.~\ref{fig:roccurves} we can see that our model has good performance ~(with $0.88$) in predicting the hierarchical relationship between part segments~(whether \textit{a} is the parent of \textit{b} or the opposite), while it is more challenging to tell whether two parts have hierarchies within the connectivity graph~(AUC $0.74$).
\begin{table}[tb]
  \centering
  \small
  \begin{tabular}{l c c}
  \toprule
     Method & $Acc_{\, edge}\, (\%)$ & $Acc_{\, obj}\, (\%)$ \\
       % & rot & trans & rot & trans  \\
    \midrule
    Random & 36.0 & 6.1 \\
    Ours &  72.7 & 31.1 \\
    \bottomrule
  \end{tabular}
      \vspace{-5pt}
    \caption{Accuracy of connectivity graph prediction. $Acc_{\, edge}$ represents the percentage of edges that are correctly recognized in $\{no\; relationship,\; parent\; of,\; child\; of\}$; $Acc_{\, obj}$ represents the percentage of objects that has correct connectivity graph with all the edges within it correctly recognized. } \label{table:ConnectPred}
    \vspace{-15pt}
\end{table}

% % \subsection{Future Application}
% \par\noindent\textbf{Future Application.}
% This section shows that apart from articulation prediction as introduced in Sec.~\ref{sec:benchmarks}, \protect\OurDataset~also enables connectivity graph prediction, which is the core component of USD format for scene representations. 
% Together with our benchmarks of articulation predictions, \protect\OurDataset~lays the foundation for 3D holistic scene understanding in the USD format by allowing training of the models that can predict both articulations and connectivity graphs, bearing great potentials in downstream applications.

\section{Implementation and Experiment Details}\label{suppsec:experiment}
This section introduces the implementation and experiment details of the experiments in Sec.~\ref{sec:method} and Sec.~\ref{sec:experiment} as well as additional results. 
\par

\subsection{Implementation Details}
We take $15\%$ of the scenes~(42 scenes) in \protect\OurDataset~ as the test set for evaluations. 
For the training and evaluation of all the methods, we downsample the point cloud with a voxel size of 2cm from the original laser scans in \cite{yeshwanthliu2023scannetpp},  in order to achieve balance between preserving geometric details and computation resources.
For the task of movable part segmentation, all points in the scene will be semantically segmented into one  of the classes $\{background, rotation, translation\}$. 
For points of $\{rotation, translation\}$, instance labels are furthered assigned for instance segmentation of movable part.
\par\noindent
\textbf{Softgroup$^\dagger$.} 
As described in the main paper, we adopt the framework of Softgroup\cite{vu2022softgroup} for this task. 
We firstly pretrain Softgroup\cite{vu2022softgroup} for 2000 epochs in the task of semantic segmentation, and then further train the semantic segmentation network together with instance grouping branch for movable part segmentation and the articulation branch for articulation parameters prediction for 220 epochs.
For interactable part segmentation, we also pretrain the network with semantic segmentation for 2000 epochs and then another 360 epochs for instance segmentation of interactable parts.
The training batch size is 6, on 2 NVIDIA A100-40g GPUs. 
Learning rate is 0.002.
\par\noindent
\textbf{Mask3D$^\dagger$.} 
For Mask3D$^\dagger$, we firstly pretrain it for semantic-instance segmentation of movable part for 1000 epochs and then further train it for joint tasks of movable part segmentation and articulation prediction for 920 epochs. 
For interactable part segmentation, we train the network for 1400 epochs.
The training batch size is 1, on 1 NVIDIA A100-40g Gpu and the learning rate is 0.0001.
In order that the input of large scenes in \protect\OurDataset~ fits in the memory of the training GPU, we randomly crop the input scenes into a $6\times6\;m^2$ cuboid during training. 
% \par\noindent
% \textbf{\protect\OurMethod~.} Similar to Mask3D$^\dagger$, we also train \protect\OurMethod~ with the semantic-instance segmentation of movable part for 1160 epochs and then further train it for joint prediction of movable parts and articulation parameters for 680 epochs. 
% For interactable part segmentation, we train \protect\OurMethod~for 640 epochs.
% The training batch size is 1, on 1 NVIDIA A100-40g GPU and the learning rate is 0.0001.
% The crop size is $6\times6\;m^2$ in cuboid during training. 

\subsection{Additional Results}
\begin{figure}
    \centering
    % First image
    \includegraphics[width=0.9\columnwidth]{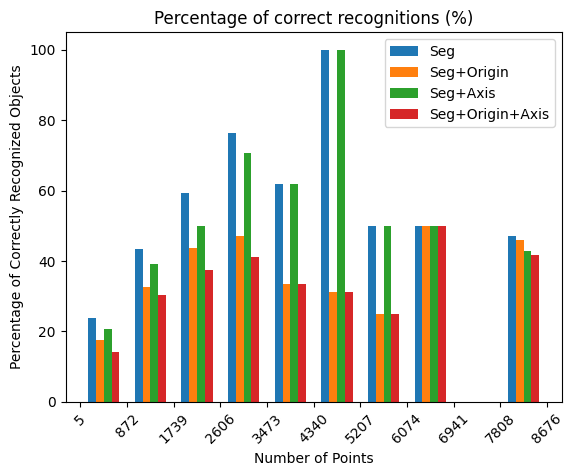}
    \vspace{-8pt}
    \caption{Percentage of correct recognitions v.s. size of movable parts}
    \label{fig:pointspercentage}
    \vspace{-8pt}
\end{figure}

In order to further understand the performance of the proposed \protect\OurMethod, we analyze its performance across objects of various sizes and categories.
In Fig.~\ref{fig:pointspercentage}, we show the performance of \protect\OurMethod~ in the task of movable part segmentation and articulation parameter prediction over the movable parts with different number of points.
From the figure we can see that for the task of movable part segmentation and articulation origin prediction, \protect\OurMethod~ performs better with middle-sized parts~(with $1739\sim5207$ points) than with the small or large parts (less than $1739$ or more than $5207$ points). 
On the other hand, articulation axis prediction is more challenging than origin prediction for the small- and middle-sized objects. 

\begin{figure}
    \centering
    % First image
    \includegraphics[width=0.85\columnwidth]{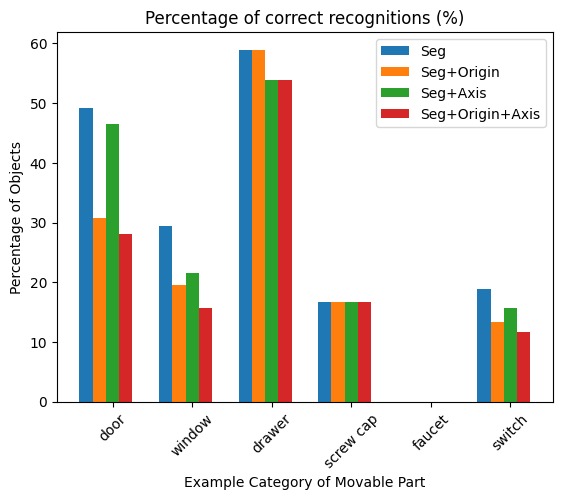}
    \vspace{-8pt}
    \caption{Percentage of correct recognitions v.s. categories of movable parts}
    \label{fig:categorypercentage}
    \vspace{-8pt}
\end{figure}
\par 
We also show the performance of \protect\OurMethod~ across the movable parts of different semantic categories. We select 6 dominant categories with significant numbers in both training and test set. 
In Fig.~\ref{fig:categorypercentage}, we can see that \protect\OurMethod~ performs better in the category of drawer~(middle sized objects) than the other categories~(small and large objects), which is in accordance with the results shown in Fig.~\ref{fig:pointspercentage}.
It is noticeable that \protect\OurMethod~ fails in the prediction of all the faucets, which is mainly because the 3D meshes of faucets are not well reconstructed and incomplete due to the complicated geometry and reflective surfaces of them. 

\section{Details of Universal Scene Description}
\label{suppsec:usd}
As mentioned in the main paper, USD organizes a scene into hierarchical entities, primitives (prims)--the building blocks representing all objects and relationships in the scene. Prims support a nested structure, where complex objects (e.g., cabinets) are represented as parent prims containing child sub-prims, modeling both individual items and grouped components of the scene.

Each prim can be assigned various attributes, e.g., position, scale, orientation, geometry, and appearance. Custom attributes can be introduced, enabling dataset-specific data to be embedded directly within the scene, such as physical properties (e.g., mass), material details, or semantic labels.

USD also supports joints that define movable connections, e.g., door hinges, as well as fixed joints. This enables the representation of both fixed and dynamic object relationships, making USD particularly suited for applications requiring detailed articulation and interaction modeling.

USD offers a highly robust, standardized format for representing complex 3D scenes, making it an ideal choice for \protect\OurDataset. It supports rich 3D data representations, including mesh geometry, semantic segmentations, connectivity and articulation definitions, and physical attributes - all within a single, unified file. It offers an efficient, lightweight alternative to datasets like MultiScan \cite{mao2022multiscan}, which require multiple scans of articulated objects in their open and closed states. USD's structure also supports non-destructive edits, facilitating scene manipulations.

USD’s relevance and utility are increasingly recognized within the research community. NVIDIA’s Isaac Sim \cite{nvidia2022isaacsim}, built on USD, is gaining popularity as the state-of-the-art simulator for research with its high-fidelity environment for realistic physics-based simulations \cite{Villasevil-RSS-24,app12178429,Jacinto,Zhou,Jiono}. USD is also widely used in research going beyond robotics simulations for projects like procedural scene generation \cite{Raistrick2024CVPR} and knowledge graph conversion for semantic querying \cite{NguyenUSD} .

\end{document}